\pdfoutput=1

\documentclass[11pt]{article}

\usepackage[preprint]{acl}

\usepackage{times}
\usepackage{latexsym}

\usepackage[T1]{fontenc}

\usepackage[utf8]{inputenc}

\usepackage{microtype}

\usepackage{inconsolata}

\usepackage{graphicx}

\usepackage{microtype}
\usepackage{hyperref}
\usepackage{url}
\usepackage{booktabs}
\usepackage{multirow}
\usepackage{graphicx}
\usepackage{subcaption}
\usepackage{wrapfig}
\usepackage{xspace}
\usepackage{amsmath}
\usepackage[nameinlink]{cleveref}

\usepackage{bbm}
\usepackage{xcolor}
\usepackage{tcolorbox}

\usepackage{arydshln}
\usepackage{colortbl}
\usepackage{soul}
\sethlcolor{lblue}
\usepackage{makecell}
\usepackage{dsfont}
\usepackage{fdsymbol}
\colorlet{lviolet}{violet!15}
\newcommand{\sodaemoji}{\raisebox{-2pt}{\includegraphics[width=1em]{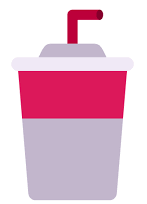}}}

\title{\sodaemoji \method: Customizable Fine-Grained Story Evaluation \\ via Chain-of-Keyword Rationalization}

\author{
Brihi Joshi$^1$\thanks{Work is mostly done at Amazon} Sriram Venkatapathy$^2$ Mohit Bansal$^2$ Nanyun Peng$^2$ Haw-Shiuan Chang$^3$\footnotemark[1]  \\
$^1$University of Southern California, $^2$Amazon AGI Foundations, \\$^3$University of Massachusets Amherst\\
\texttt{brihijos@usc.edu} \\
\texttt{\{vesriram,mobansal,pengnany\}@amazon.com}\\
\texttt{hschang@cics.umass.edu}
}

\newcommand{\method}{\textsc{CoKe}\xspace}

\begin{document}

\maketitle

\begin{abstract}

Evaluating creative text such as human-written stories using language models has always been a challenging task -- owing to the subjectivity of multi-annotator ratings. 
To mimic the thinking process of humans, chain of thought~\citep{wei2023chainofthought} (CoT) generates free-text explanations that help guide a model's predictions and 
Self-Consistency~\citep{wang2022self} (SC) marginalizes predictions over multiple generated explanations. 
In this study, we discover that the widely-used self-consistency reasoning methods cause suboptimal results due to an objective mismatch between generating `fluent-looking' explanations vs. actually leading to a good rating prediction for an aspect of a story. 
To overcome this challenge, we propose \textbf{C}hain-\textbf{o}f-\textbf{Ke}ywords (\method), that generates a sequence of keywords \textit{before} generating a free-text rationale, that guide the rating prediction of our evaluation language model. 
Then, we generate a diverse set of such keywords, and aggregate the scores corresponding to these generations. 
On the StoryER dataset, \method based on our small fine-tuned evaluation models not only reach human-level performance and significantly outperform GPT-4 with a 2x boost in correlation with human annotators, but also requires drastically less \# of parameters.



\end{abstract}



\section{Introduction}

Evaluating stories is an important and time-consuming job for professionals in the entertainment industry. For example, novel competition judges, book editors, or movie producers might need to select the best story from thousands of submissions according to their tastes and the understanding of the market.

As LLMs get better at judging story quality, automatically evaluating human-written stories becomes practical. However, there are still several challenges to overcome. First, judgements from off-the-shelf LLMs might be biased towards the preference of particular annotators during the alignment stage, which could be very different from the tastes of the desired population. Second, humans are extremely subjective in judging creative writing like stories, which is often demonstrated in their creativity:
Some readers or professional reviewers would think character shaping is the most critical component for evaluating a story, whereas others might like or dislike the characters along with some other components, like the scene description mentioned in the story.
This lack of consensus in likes and dislikes, along with differences across aspects (e.g. character shaping, ending, etc) in the story makes evaluating human-written stories an extremely difficult task.

The desired human evaluation here would entail that we collect diverse opinions from different readers/reviewers to estimate a \textit{average} opinion of the story from a desired population, but this is extremely tedious and expensive.
This high cost has motivated automatic measures for evaluating the stories written by humans.
In this study, we aim at building an automatic story evaluation system that can 1) provide fine-grained  evaluation for a human-written story in predefined and/or customized aspects, 2) provide a set of \textit{rationales} that model diverse opinions of multiple humans and help us better predict the average score for different aspects of the story, and 3) be easily customized toward the opinions of the desired population (i.e., fine-tunable using the collected human judgements and explanation).

\begin{figure*}
    \centering
    \includegraphics[width=\linewidth]{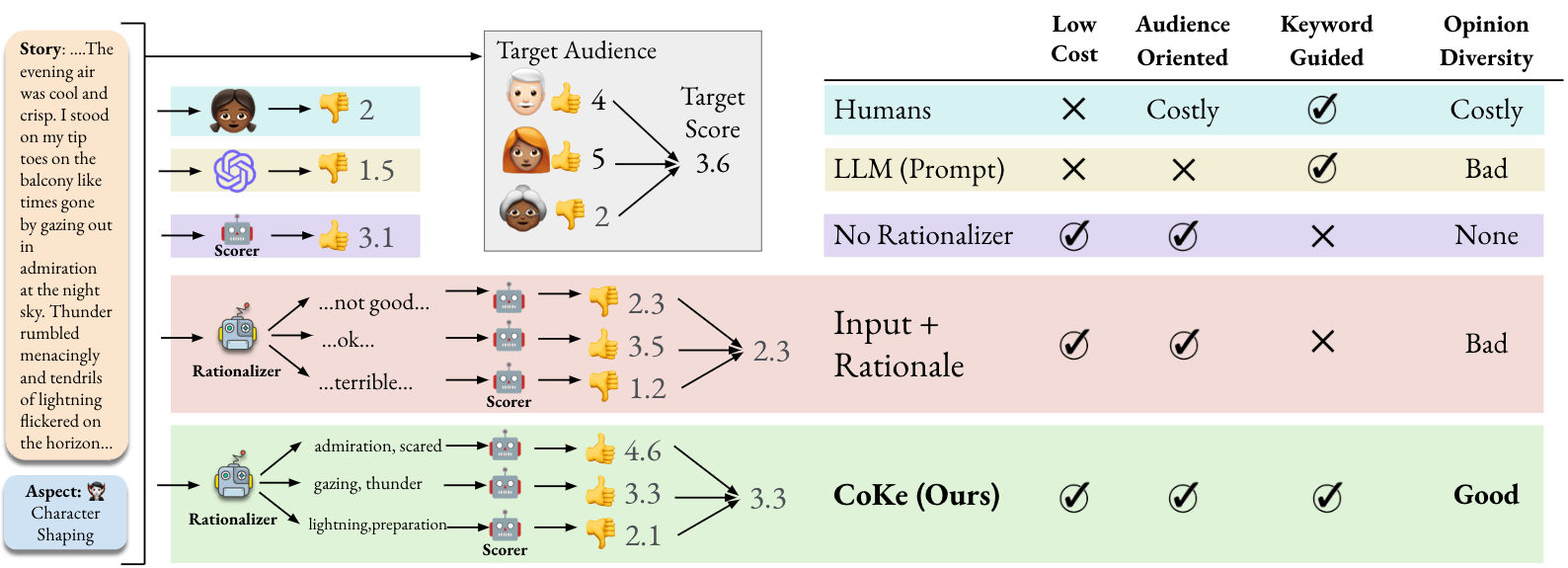}
    \caption{\method provides a low-cost, audience-oriented (customizable), and keyword-guided approach to evaluating stories by generating and scoring diverse keyword sequences that explain a fine-grained aspect-story pair. 
    }
    \label{fig:pitch_figure}
\end{figure*}

The reason-then-predict approaches like Chain of Thought (CoT) \citep{wei2023chainofthought} not only improve the interpretability of the said predictions by generating rationales but also improve downstream performance in predictions \citep{wei2023chainofthought,wang-etal-2023-scott}.
Using these approaches, Large Language Models (LLMs) can score arbitrary aspects of a story without any additional training.
However, for story evaluation particularly, the scores from prompting LLMs might deviate from the population average of our target audience, along with significant cost induced by large token lengths of such inputs.

Fine-tuning a small Language Model (LM) to directly predict the population average of annotators is a cheap viable alternative, but does not provide rationales while also being inflexible w.r.t. how granular we want the story to be evaluated (e.g., character shaping of the vampire, ending w.r.t. a certain character, etc). Another option is fine-tuning a small LM to generate free-text rationals for CoTs and use the self-consistency \citep{wang2022self} approaches to marginalize over multiple sampled CoTs. However, we discover that the free-text rationals tend to reduce the diversity of CoTs' rating predictions and deviate the average prediction rating from the population average.


In order to mitigate this shortcoming we propose Chain-of-Keywords, \method, which consists of two simple yet effective modifications to regular CoT approaches.
First, instead of just generating a free-text rationale, we generate a chain of \textit{keywords} before generating a rationale that can describe salient concepts in and outside the story.
Our intuition is that keywords help prevent the learning and generation of annotator artifacts (like sentiment-laden words and other personal descriptors like `I think, I feel', etc), which assists with the objective misalignment we see in CoT approaches.
Like SC, instead of generating one rationale, it samples \textit{multiple} keyword rationales, which simulates annotator diversity and helps better estimate the population average.
Therefore, \method uses the generated keywords to score a story, and the corresponding generated rationale for interpreting the story, as shown in \Cref{fig:pitch_figure}.

On StoryER~\citep{chen-etal-2022-storyer}, a fine-grained story evaluation benchmark \citep{chen-etal-2022-storyer}, we show that \method can better estimate population averages as compared to LLM baselines using GPT-3.5 (text-davinci-003) and GPT-4 (gpt-4-0613) \citep{brown2020language, ouyang2022training}, as well as open-source LLMs like LLaMa-2-7B-Chat \citep{touvron2023llama} and Mistral-7B-Instruct \citep{jiang2023mistral}.
We also show that \method consistently outperforms self-consistency and approaches based on supervised fine-tuning, including those where the rationale generated is specifically aligned to that of annotator-written explanations using reinforcement learning (RL), as well as improved correlations on human evaluations as compared to baselines.
Furthermore, we also show that \method can work effectively even when built on smaller LMs as its backbone (approx. 58x fewer \# of parameters than GPT-3.5), while surpassing GPT-3.5 by 2.18x improvement in correlation metrics with the target annotator population.
To the best of our knowledge, \method is a first rationalize-then-predict approach for fine-grained story evaluation surpassing LLM performance for this task, and reaches human-level performance in the StoryER dataset \citep{chen-etal-2022-storyer}.\footnote{Our code and models will be released.}

\section{Problem Formulation}
\label{sec:formulation}
We begin by describing our task setup and why the task is challenging.
\paragraph{Task setup.} 
We are given a story, along with an aspect, with respect to which we want to evaluate the story. 
The aspect can focus on certain semantic or literary features of the story \citep{gulich1986story}, like \textit{humor}, \textit{character shaping}, etc.
Our task is to evaluate the story with respect to the given aspect and provide a Likert rating between $1$ and $5$, where a higher score implies the story is better with respect to the aspect. 

We assume that there exists a dataset that consists of the story-aspect ratings and the explanations for the ratings. 
One story-aspect pair could be annotated by multiple annotators from our target audience. 
Any automated story evaluation system should provide a single score for an aspect-story pair that is close to the average ratings from annotators, without modeling the individual annotator~\citep{sap2021annotators,wang2023learning}. 


\begin{figure}
  \begin{center}
    \includegraphics[width=0.45\textwidth]{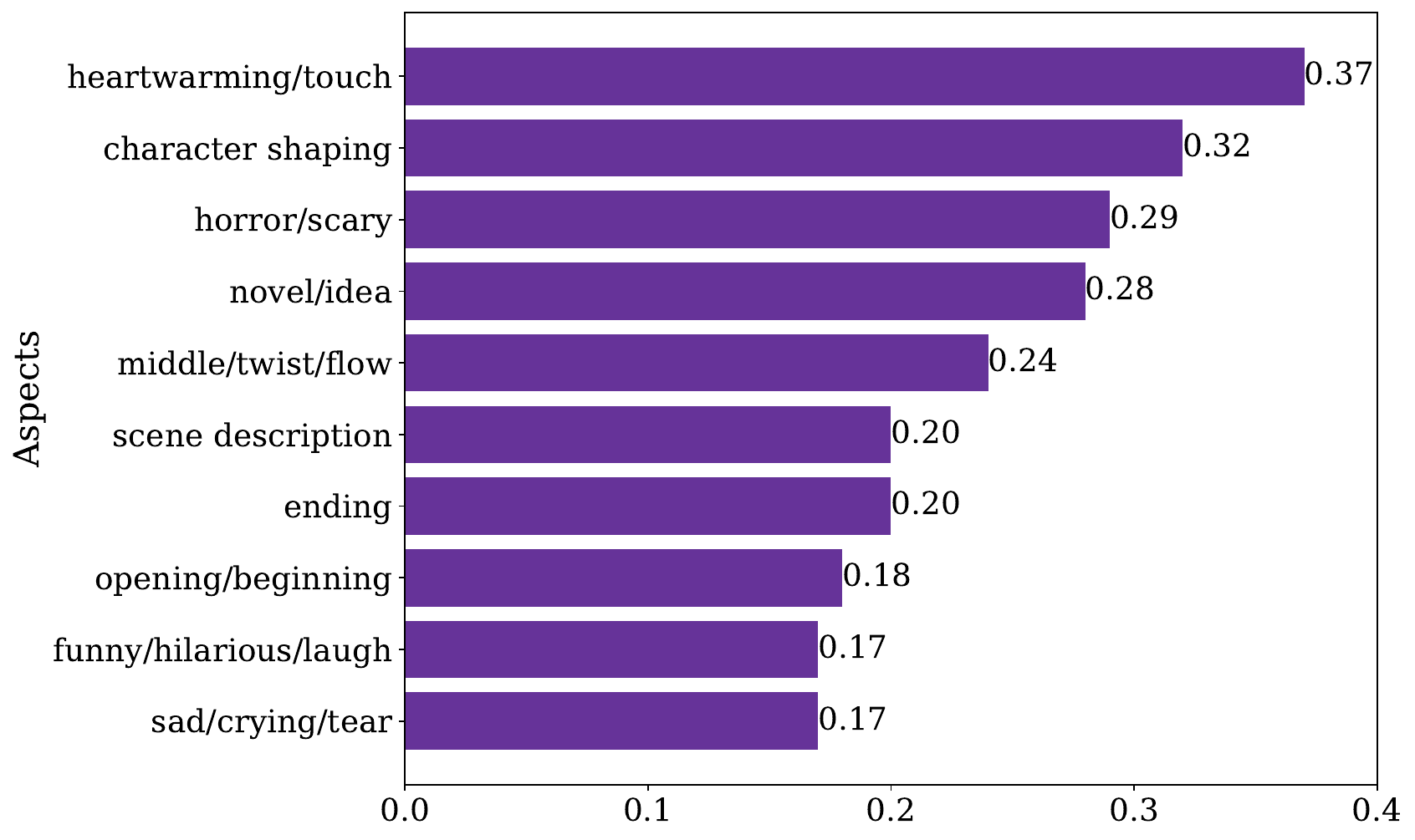}
  \end{center}
  \caption{ICC annotator agreements scores for the stories with a certain aspect in the training set.}
  \label{fig:annotator_agreements}
\end{figure}

\paragraph{Story evaluation is an extremely subjective task.}
We use the StoryER dataset \citep{chen-etal-2022-storyer} for our task.
What is interesting to note here is even though all annotators have to focus on a certain aspect of the story, human ratings are still extremely subjective.
In the StoryER dataset, we calculate Intraclass Correlation Coefficient (ICC) scores \citep{iccscore} to evaluate annotator agreements within annotators for a given aspect, across all the possible stories which are marked with that aspect (\autoref{fig:annotator_agreements}).
The `heartwarming' aspect has the highest agreement of $0.37$, which is still considered to be poor while interpreting ICC scores \citep{iccscore}.


\paragraph{Limitation of CoTs for story evaluation.}
\label{sec:subjectivity}


Self-consistency \citep{wang2022self} is an approach that extends Chain of Thought (CoT) \citep{wei2023chainofthought} to capture the diverse opinions of humans. 
\citet{wang2022self} sample various free-text rationales and marginalize the different predictions based on the generated CoT. However, it is very difficult to decode all possible rationales.
Furthermore, there could be some objective misalignments between generating highly probable and \textit{coherent} rationales and predicting the final ratings from annotators~\citep{jia-etal-2020-mitigating}. For example, let's say in our training data, our vampire stories and their corresponding explanations are all good and positive. Then, if there are some vampire stories that are boring and contain some grammatical errors during the testing test time, the LM does not know how to generate a negative rationale for a vampire story, so it is forced to generate coherent but biased rationales, which lead to positive rating predictions.

\section{Chain-of-Keywords (\method)}
\label{sec:method}

\begin{figure*}
    \vspace{-0.5cm}
    \centering
    \includegraphics[width=0.97\linewidth]{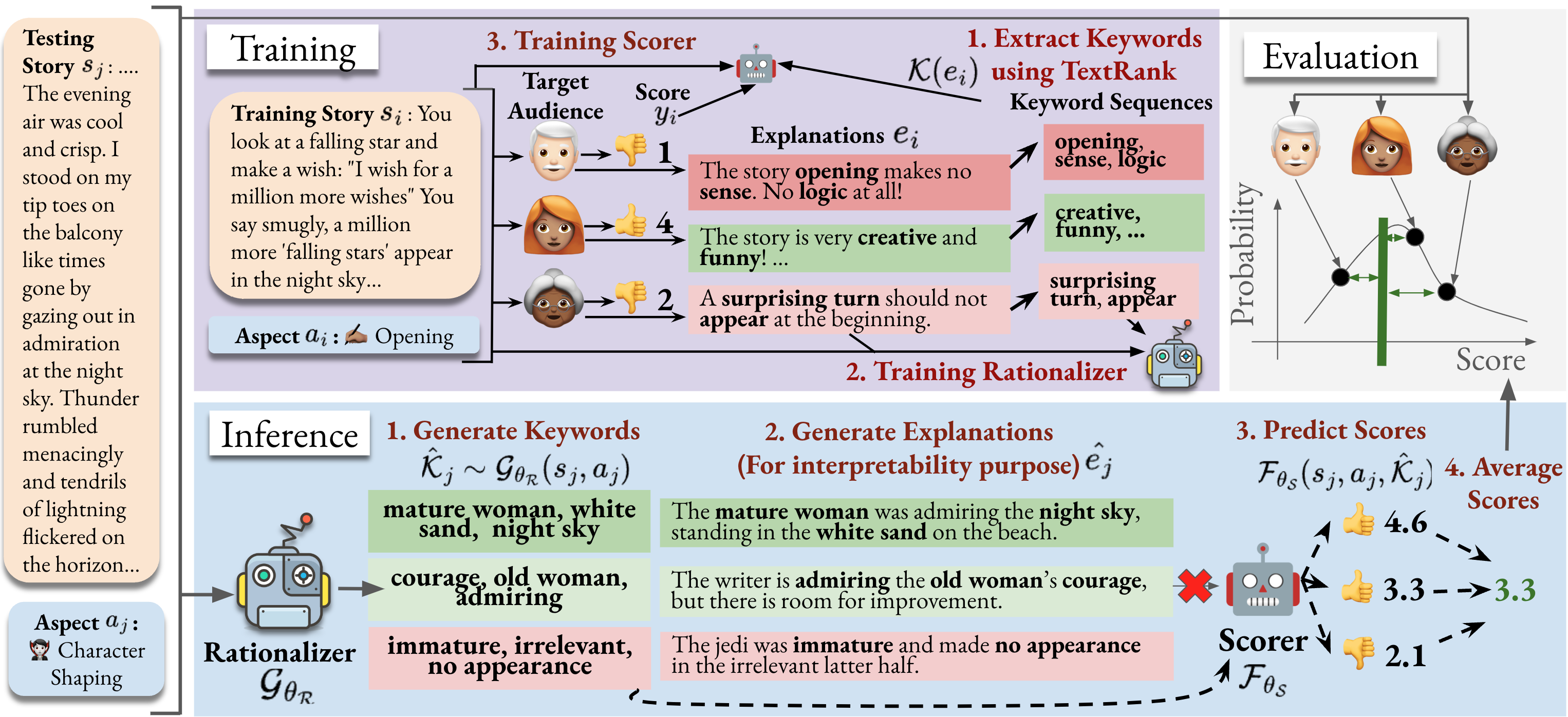}
    \caption{During training, \method extracts keywords from annotator explanations and train rationalizers and scorers. During inference, \method first samples candidate keyword sequences (for the scorer) and explanations (for better interpretability), and then score the individual generated candidates before aggregating them. Our purpose is to obtain a better \textit{population average} that can capture diverse annotator scores.}
    \label{fig:method}
\end{figure*}




There are three kinds of words in a free-text explanation: sentiment words, keywords referring to the concepts in the story, and the functional words (e.g., stop words). 
We view the sentiment and functional words as an artifact for story evaluation because they only provide the information that the rating has already provided and could induce a bias in CoT's rating prediction. 
This is because the probability of generating a positive sentiment word might be affected more by the nearby function words than by the quality of the input story and thus, the positive sentiment in the explanation would heavily bias the CoT to predict a high score.

For example, we observe that most positive rationales in the StoryER dataset are much more likely to contain ``\textit{I}'' while the most negative rationales have much more ``\textit{It}''. In the positive rationales, \textit{I} is the 8th likely words (1.8\%) while \textit{It} is the 14th likely words (0.6\%). In the negative rationales, \textit{I} is the 16th likely words (0.9\%) while \textit{It} is the 7th likely words (1.5\%). If we observe some rationales starting with ``\textit{I like}'' or ``\textit{I love}'' in the training vampire stories, ``\textit{I}'' could become the most likely first word in the generated rationale for a bad testing vampire story, which bias the CoT to output \textit{like/love} and a high rating at the end.







We leverage these intuitions to build \method in the following manner (shown in \Cref{fig:method}).
First, a language model is fine-tuned to generate \textit{keywords}, along with a free-text explanation conditioned on those keywords, that inspects the story w.r.t the aspect.
These keywords are in the form of phrases (from the story itself) that specifically do not contain artifacts. 
From this language model's decoder, we sample multiple keyword sequences, intended to simulate diverse annotator opinions.
A trained scorer model is then used to produce score predictions from aspect-story-keyword triples, and scores for all individual candidate keyword sequences are averaged to produce the final score.



More concretely, let $\mathcal{D}_{Tr}$ and $\mathcal{D}_{Te}$ be the training and test datasets respectively. They are composed of the story-aspect-explanation-rating tuple $(s_i, a_i, e_i, y_i)$. 
For example, in StoryER, $s_i$ is a human-written story from WritingPrompts \citep{fan-etal-2018-hierarchical}, $a_i$ is one of the predefined aspects, $y_i$ is the rating from an annotator, and $e_i$ is the text justification for $y_i$. 
If two annotators label the same story and aspect, the $s_i$ and $a_i$ would be the same for the two tuples.

\method consists of two components: a rationalizer model, $\theta_R$, and a scorer model, $\theta_S$. The rationalizer is a seq2seq language model that is fine-tuned to generate rationales, given an aspect-story pair as an input: $\hat{\mathcal{K}_j} \sim \mathcal{G_{\theta_R}}(s_j, a_j)$, while the scorer is a regression language model that is fine-tuned to predict a floating point score, given aspect-story-rationale triplets as an input: $y_j = \mathcal{F_{\theta_S}}(s_j, a_j, \hat{\mathcal{K}_j})$. We detail the training and inference process of \method below and further conduct ablations on different components of \method to justify our keyword extraction step and other design decisions in \Cref{sec:ablations}.




\paragraph{Training in \method.} 

Given story-aspect-explanation-rating tuple $(s_i, a_i, e_i, y_i)$, we first extract the \textit{keywords} from the annotator-written explanation $e_i$ and train our rationalizer to first generate the extracted keyword sequence $\mathcal{K}(e_i)$ before generating the explanation $e_i$. We template the inputs for the rationalizer to contain both the aspect and story -  \texttt{aspect: <aspect> story: <story>}, and the output is a chain of keywords, followed by a free-text explanation that is conditioned on the keywords, which looks like - \texttt{keywords: <key1, key2, $\ldots$, keyn> rationale: <natural language explanation>}.

For the scorer, we provide the story $s_i$, aspect $a_i$, and extracted keyword sequence $\mathcal{K}(e_i)$ as the input and ask it to predict the rating from the annotator $y_i$. The input to the model looks like - \texttt{aspect: <aspect> story: <story> keywords: <keywords>} and the loss function is the mean squared error.

\begin{table*}[!h]
\centering
\scalebox{0.72}{
\begin{tabular}{ccccccc}
\toprule
\multirow{2}{*}{\textbf{Setting}}& \textbf{Rationale} & \multirow{2}{*}{\textbf{Rationalizer}} & \multirow{2}{*}{\textbf{Scorer}} &\multicolumn{3}{c}{\textbf{Metrics}} \\
\cline{5-7}
  & \textbf{for Scorer} &  &  & \textbf{Pearson's $\rho$ ($\uparrow$)} &\textbf{ MSE ($\downarrow$)} & \textbf{F1-Score ($\uparrow$)}\\
\midrule
\midrule
\multirow{12}{*}{\textbf{LLM}} & - &  None & GPT-3.5  & 0.0240 & 0.5172  & 0.2277 \\
& - & None & GPT-3.5 5-shot  & 0.1440 & 0.2703 & 0.4751 \\
& Explanation & \multicolumn{2}{c}{GPT-3.5 CoT} & 0.1049 & 0.2290  & 0.4833 \\
& Explanation & \multicolumn{2}{c}{GPT-3.5 CoT SC Mean} & 0.1303  & 0.1970 & 0.5267    \\
& Explanation & \multicolumn{2}{c}{GPT-4 CoT} & 0.1093 & 0.3039 & 0.4199   \\
& Explanation & \multicolumn{2}{c}{Mistral-7B-Instruct CoT} & 0.0573  & 0.5113  & 0.3760 \\
& Explanation & \multicolumn{2}{c}{Mistral-7B-Instruct CoT 5-shot} & 0.0596  & 0.5019 &	0.3760   \\
& Explanation & \multicolumn{2}{c}{Mistral-7B-Instruct CoT-SC MV} & 0.0648  & 0.5252 & 0.3760\\
& Explanation & \multicolumn{2}{c}{Mistral-7B-Instruct CoT-SC Mean} & 0.1023 & 0.4998 & 0.3740 \\
& Explanation & \multicolumn{2}{c}{Mistral-7B-Instruct CoT-SC Mean 5-shot} & 0.1266 & 0.4578 & 0.3940  \\
& Keywords & \multicolumn{2}{c}{Mistral-7B-Instruct CoT } & 0.0277 & 0.6892 & 0.2007  \\
& Keywords & \multicolumn{2}{c}{Mistral-7B-Instruct CoT 5-shot} & 0.0300 &	0.6676 &	0.2101  \\
\midrule
\multirow{5}{*}{ \shortstack[l]{\textbf{Supervised} \\ \textbf{Fine-tuning}}} & Explanation & T5-Small & DeBERTa-V3-Small  & 0.0904 & 0.1339  & 0.5827 \\
& Explanation & T5-Small PPO & DeBERTa-V3-Small  & 0.0779 & 0.1118 & 0.5773 \\
& Explanation & \multicolumn{2}{c}{T5-Small CoT}  & 0.0676  & 0.1698 & 0.5622 \\
& - & None  & T5-Small & 0.0712  & 0.1620 & 0.5647 \\
& - & None  & T5-Small Prob-avg & 0.2451 & 0.1331 & 0.6162 \\
\midrule
\rowcolor{gray!30} \textbf{Human}& Explanation & \multicolumn{2}{c}{Human}  & 0.3037 & 0.1972 & 0.4998 \\
\rowcolor{lviolet}  & \cellcolor{lviolet}Keywords & \cellcolor{lviolet}T5-Small & \cellcolor{lviolet}DeBERTa-V3-Small & \cellcolor{lviolet}\textbf{0.2900}  & \cellcolor{lviolet}\textbf{0.0912} & \cellcolor{lviolet}\textbf{0.6334}\\
\cellcolor{lviolet} \multirow{-2}{*}\method & \cellcolor{lviolet}Keywords & \cellcolor{lviolet}T5-3B  & \cellcolor{lviolet}DeBERTa-V3-Small & \cellcolor{lviolet}\textbf{0.3142}  & \cellcolor{lviolet}\textbf{0.0811} & \cellcolor{lviolet}\textbf{0.6509} \\
\bottomrule
\end{tabular}
}
\caption{We compare \method to other baselines that use rationalize-then-predict paradigms in StoryER. For all Self-Consistency (SC) variations, we average over 40 samples as done by \cite{wang2022self}. For \method, we provide the best performing setting with $\mathcal{N}=100$ samples.
}
\label{tab:main_baseline_results}
\end{table*}

\paragraph{Inference in \method.} 
After training $\theta_R$ and $\theta_S$ separately, \method inference is explained below.

We simulate diversity in annotators by sampling multiple candidate keyword sequences using $\mathcal{G_{\theta_R}}$, and then marginalize the candidate rationales by taking a mean over scores of individual candidates. 
This score is represented as follows - 
\begin{equation}
    \mathbb{E}_{\hat{\mathcal{K}}_j \sim \mathcal{G_{\theta_R}}(s_j, a_j)} \left[\mathcal{F_{\theta_S}}(s_j, a_j, \hat{\mathcal{K}}_j)\right],
\end{equation}
where $(s_j, a_j)$ is a testing example from $\mathcal{D}_{Te}$. 
 
Since finding all possible $\hat{\mathcal{K}}_j$ is not feasible to calculate the expectation term, we conduct Monte Carlo simulations over a set number of samples, $\mathcal{N}$, over which we average the score. Notice that $\mathcal{G_{\theta_R}}$ could also generate the free-text explanations, $\hat{e}_j$, after the keywords, but they are just for interpretability purpose and won't affect the final score prediction. 





\section{Experiments}
In this section, we evaluate \method, LLMs with sophisticated inference strategies, supervised fine-tuning, along with \method ablations.

\subsection{Evaluation Setup}

We train our T5 \citep{raffel2023exploring} rationalizer and DeBERTa-V3 \citep{he2021debertav3} scorer using the training set of StoryER~\citep{chen-etal-2022-storyer} dataset and evaluate \method using its official test set. 
We first filter out story-aspects pairs that are only rated by one annotator and normalize the scores from annotators and models into the range from 0 to 1, using min-max normalization where max=5 and min=1. 
Given an input story-aspect pair, each model can only produce a single score. 
As shown in the evaluation block of \Cref{fig:method}, we compare the output score with each annotator-provided score separately and the prediction that is closer to the average of all the human scores would perform better. 
This procedure allows us to compare each model with human performance and handle the varying numbers of human annotators, given the same input pair in StoryER.

We report three metrics for every evaluation conducted -- Pearson's Correlation Coefficient ($\rho$), Mean Squared Error (MSE), and F1-score on binarized score values, thresholded using a value of 0.5. 
We use the Pearson correlation coefficient as the main metric because the global score average might be very different for different human annotators or different models. 
For example, the GPT-4's scores are found to be over-generous sometimes~\citep{doostmohammadi2024reliable, gmyrek2024technological}.

\subsection{Human vs. \method}

To estimate human performance, we use one annotator as the \textit{prediction} that is compared to the other annotators for each pair of story and aspect.
This process is repeated for every annotator's rating and story-aspect pair.
In \Cref{tab:main_baseline_results}, we see that \method's best configuration significantly outperforms the human performance in MSE and slightly in Pearson's $\rho$, which shows that \method's prediction is closer to the population average than the individual human.




\subsection{LLMs vs. \method}
\label{sec:llm}

We prompt a mix of closed and open-sourced Large Language Models like \textbf{GPT-3.5} (text-davinci-003) and \textbf{GPT-4} (gpt-4-0613) \citep{brown2020language,ouyang2022training,openai2024gpt4}, 
and \textbf{Mistral 7B Instruct} \citep{jiang2023mistral} to generate a score for a given story-aspect pair. 
These models can be prompted to generate a score as-is or with a rationale, with the help of Chain of Thought (\textsc{CoT}) prompting \citep{wei2023chainofthought}.
We evaluate zero- and few-shot prompting \textit{without} CoT and \textit{with} CoT.
As seen in \Cref{tab:main_baseline_results}, our approach always outperforms strong LLMs prompted with CoT prompts to score an aspect-story pair.
We can note a 3x improvement in Pearson's $\rho$ shown by \method ($\approx$3B) in comparison to GPT-3.5 CoT while having an estimated 58x lesser number of parameters that GPT-3.5 ($\approx$175B).


We also run Self-Consistency (SC) approaches as shown by \cite{wang2022self}.
We generate 40 CoT predictions per story-aspect pair in the test set and show two variations to aggregate scores provided by these CoTs: \textbf{Majority Voting} (MV)
and \textbf{Mean}, a more suitable method for story evaluation tasks.
\Cref{tab:main_baseline_results} shows that \method correlates with the population averages better than the SC approaches. \Cref{appendix:diversity} further demonstrates that \method also outputs much more diverse ratings than SC.



\subsection{Supervised Fine-tuning (SFT) vs. \method}

Rationalization approaches pre-dating LLMs also fine-tuned smaller LMs to generate rationales, and then predict an answer based on the rationale and the input \citep{wiegreffe-etal-2021-measuring,marasović2022fewshot}. The approaches are cost-efficient and could be easily customized for the target audience.
We use the \textit{pipeline approach} \citep{wiegreffe-etal-2021-measuring} for generating both the rationales and scores for a given aspect-story pair (\textbf{T5-small + DeBERTa-V3-Small}). 
The \textit{pipeline} is the same as \method except that T5 generates only one free-text explanation rather than multiple keyword sequences (i.e., $\mathcal{N}=1$ and $\mathcal{K}(\cdot) = \mathbbm{1}(\cdot)$).

A shortcoming of the \textit{pipeline} approaches is that they do not focus on the quality of the rationales that are generated.
To mitigate the explanation distribution mismatch \citep{kirk2024understanding} between annotators and generation, 
we added an additional alignment step, where generated rationales would be compared to the annotator-provided explanations using a Cider score reward \citep{vedantam2015cider}, and used as feedback into the \textsc{Rationalizer} using the PPO algorithm \citep{schulman2017proximal, Ramamurthy2022IsRL} (\textbf{T5-small PPO + DeBERTa-V3-Small}).
Surprisingly, in \Cref{tab:main_baseline_results} we see that specifically aligning generations with annotated explanations does not aid downstream scoring performance.
This validates that explicitly improving rationale quality does not improve downstream aspect-story evaluation~\citep{kirk2024understanding,florian2024exploring}.


In another approach, we fine-tune a T5 model to first generate an explanation, followed by a score (\textbf{T5-small CoT}) \citep{kim-etal-2023-cot} without training another scorer model.
\Cref{tab:main_baseline_results} shows that SFT approaches are not at par with LLM-based baselines, and thus by default, lag behind \method.
Based on \citet{marasović2022fewshot}, we also make a modification to SFT-CoT, where instead of generating a score conditioned on the explanation, we generate the score before generating the explanation (\textbf{T5-small}) \citep{marasović2022fewshot}.
Instead of sampling score, we also calculate \textit{expected} predicted score for which we compute the weighted average according to the probabilities of each score token (\textbf{T5-small Prob-avg}).
This leads to significant improvements in Pearson's $\rho$ over other SFT approaches in \Cref{tab:main_baseline_results}, which shows the importance of generation diversity in this task.



\subsection{\method Ablations}
\label{sec:ablations}

\begin{table}
\centering
\scalebox{0.51}{
\begin{tabular}{llc}
\toprule
& & \multicolumn{1}{c}{Metrics} \\
\cmidrule{3-3}
Rationalizer & Scorer & Pearson
\\
\midrule
\midrule
- &  $(s, a) \rightarrow$ DeBERTa-V3 Small & 0.2718  \\
- &  $(s, a) \rightarrow$ DeBERTa-V3 Large & 0.2697\\
\midrule
T5 Small $\rightarrow (e)$ & $(s, a, e) \rightarrow$ DeBERTa-V3 Small & 0.2040 \\
T5 Small $\rightarrow (e)$ & $(a, e) \rightarrow$ DeBERTa-V3 Small & 0.1912\\
T5 Small $\rightarrow (\mathcal{K}_{\text{TF-IDF}}(e))$ & $(s, a, \mathcal{K}_{\text{TF-IDF}}(e)) \rightarrow$ DeBERTa-V3 Small & 0.2548 \\
T5 Small $\rightarrow (\mathcal{K}_{\text{Rake}}(e))$ & $(s, a, \mathcal{K}_{\text{Rake}}(e)) \rightarrow$ DeBERTa-V3 Small & 0.2081 \\
T5 Small $\rightarrow (\mathcal{K}_{\text{TextRank}}(e))$ & $(s, a, \mathcal{K}_{\text{TextRank}}(e)) \rightarrow$ DeBERTa-V3 Small & 0.2727\\
T5 Small $\rightarrow (\mathcal{K}_{\text{TextRank}}(e))$ & $(a, \mathcal{K}_{\text{TextRank}}(e)) \rightarrow$ DeBERTa-V3 Small & 0.1924\\
\midrule
\rowcolor{lviolet} T5 Small $\rightarrow (\mathcal{K}_{\text{TextRank}}(e), e)$ & $(s, a, \mathcal{K}_{\text{TextRank}}(e)) \rightarrow$ DeBERTa-V3 Small & 0.2800 \\
\rowcolor{lviolet} T5 Large $\rightarrow (\mathcal{K}_{\text{TextRank}}(e), e)$ & $(s, a, \mathcal{K}_{\text{TextRank}}(e)) \rightarrow$ DeBERTa-V3 Small & 0.2834 \\
\rowcolor{lviolet} T5 3B $\rightarrow (\mathcal{K}_{\text{TextRank}}(e), e)$ & $(s, a, \mathcal{K}_{\text{TextRank}}(e)) \rightarrow$ DeBERTa-V3 Small & 0.2887\\
\rowcolor{lviolet} T5 3B $\rightarrow (\mathcal{K}_{\text{TextRank}}(e), e)$, $\mathcal{N}=100$ & $(s, a, \mathcal{K}_{\text{TextRank}}(e)) \rightarrow$ DeBERTa-V3 Small & \textbf{0.3142}\\
\bottomrule
\end{tabular}
}
\caption{Ablation study. $s$ is a story, $a$ is an aspect, $e$ is an explanation, and $\mathcal{K}(.)$ is a keyword extraction function. For rationalizers, $\mathcal{N}=10$ except for the last row. \method (Ours) in the last four rows are highlighted.
}
\label{tab:ablation}
\end{table}

\paragraph{No Rationalizer in \method. }

During inference, \method's scorer takes in the aspect-story pair, along with the generated keywords from a fine-tuned rationalizer model. Here, we remove the rationales from the input of the scorer and fine-tune DeBERTa-V3 models to predict a score only based on the aspect-story pair (s,a). In \Cref{tab:ablation}, we see that the $\mathbf{(s, a) \rightarrow}$ \textbf{DeBERTa-V3 Small/Large} baselines are strong, surpassing performances by LLMs in \Cref{tab:main_baseline_results}, while being significantly worse than \method. Furthermore, it cannot provide rationales or consider the user-specified aspects/keywords.





\begin{figure}
  \begin{center}
    \includegraphics[width=0.48\textwidth]{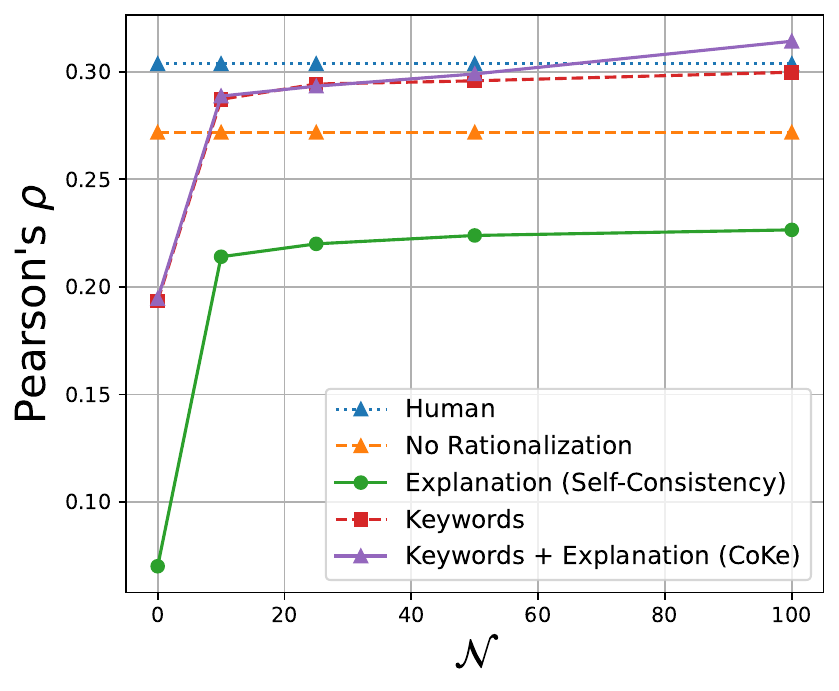}
  \end{center}
  \vspace{-0.2cm}
  \caption{Pearson's $\rho$ increases with the larger number of candidate generations ($\mathcal{N}$) in \method and it's ablations. The rationalizer model here is T5-3b. We note that increasing the diversity of generation helps with better estimation of population preferences.}
  \label{fig:ablation_n}
\end{figure}

\paragraph{Varying Rationales in \method. }

In \Cref{sec:method}, we use $\mathcal{K}(\cdot)$ to 
extract keywords from the gold explanations $e$ in the dataset (during training of the rationalizer and scorer). First, we remove the keyword extraction step $\mathcal{K}(\cdot)$ in the baseline $\mathbf{(s, a, e) \rightarrow}$ \textbf{DeBERTa-V3 Small} to verify the design. This is equivalent to \textbf{T5-small + DeBERTa-V3-Small} in \Cref{tab:main_baseline_results}, except that we use $\mathcal{N}=10$ rather than $\mathcal{N}=100$ here. Its $\rho$ (0.2040) is much worse than the $\rho$ of $\mathbf{(s, a) \rightarrow}$ \textbf{DeBERTa-V3 Small} (0.2718). To investigate the reason, we conduct another baseline that removes the story, the most important signal, from the input of the scorer ($\mathbf{(a, e) \rightarrow}$ \textbf{DeBERTa-V3 Small}) and we find that its $\rho$ only degrades slightly to 0.1912. This indicates that the scorer relies too much on the signal in explanation (e.g., sentiment words) to predict the ratings and ignore the signal in the story itself.





We also try different keyword extractors: TF-IDF \citep{frank1999domain}, Rake \citep{rose2010automatic} and TextRank \citep{mihalcea2004textrank}.
After keyword extraction, we remove all sentiment words from the keyword sequence. In \method, we use TextRank for our choice of $\mathcal{K}(\cdot)$ due to its best performance in \Cref{tab:ablation}.

Finally, we find \textbf{T5 Small} $\mathbf{\rightarrow (\mathcal{K}_{\text{TextRank}}(e), e)}$ in \method (0.2800) slightly outperforms \textbf{T5 Small} $\mathbf{\rightarrow (\mathcal{K}_{\text{TextRank}}(e))}$ (0.2727), which implies that predict the free-text explanations after keywords further improves predictions of the scorer, even though the scorer does not consider the generated explanations during inference time. Furthermore, the coherent free-text explanations could also improve the interpretability of the predicted ratings (see examples in \Cref{tab:apx:keyword_examples}). 

\paragraph{Rationalizer Sizes in \method. }

In \Cref{tab:ablation}, we also show how scaling the size of the rationalizer helps improve Pearson's $\rho$. 
We note that our best-performing setup includes a T5 3B model as the rationalizer, along with the DeBERTa-V3-Small model as a scorer.
It is interesting to note that \method ends up being 2.18x better than GPT-3.5 in \Cref{tab:main_baseline_results} while being approximately 58x smaller in parameter size as compared to it.


\paragraph{Varying $\mathcal{N}$ in \method. }

In \Cref{fig:ablation_n}, we also compare varying the number of candidate generations from $\mathcal{G_{\theta_R}}$ while scoring an aspect-story pair.
We see that increasing the number of generations, $\mathcal{N}$ improves the Pearson's Correlation Coefficient, thereby supporting our hypothesis that diversity of generations can help mimic various annotator preferences.
Increasing $\mathcal{N}$ for \method helps it surpass the human performance.
We also note that increasing $\mathcal{N}$ is less costly as compared to LLM approaches shown in \Cref{tab:main_baseline_results}, because \method uses a smaller, finetuned LM.

\section{Applications of Keywords in \method}

The keyword rationales generated by \method not only significantly improve the performance, but also being faithful because they are used as input for the scorer, similar to other faithful rationalization approaches like \citet{jain-etal-2020-learning}. Moreover, the keywords provide more interpretable evaluation and more fine-grained evaluation based on user-provided keywords.


\subsection{Human Evaluation for Considering User-provided Keywords}

To support our results further, we conduct a small human evaluation experiment.
For this task, we ask two annotators each to first read the story and the corresponding aspect and ask them to provide \textit{one keyword or keyphrase} of their choice, along with a score that helps them to evaluate aspect-story-keyword triple (\Cref{sec:apx:human_eval}).
We conduct this experiment on a subset of 100 story-aspect pairs from our test set, with the help of annotators recruited via Amazon Mechanical Turk\footnote{\url{https://www.mturk.com/}}.
Here, we compare \method with the No Rationalization baseline and find that \method utilizes the keyword provided by the annotators and leads to an $29.2\%$ relative improvement over the Pearson's Correlation Coefficient score.
This validates that \method can better correlate with annotator-provided fine-grained keywords that baselines that do not have any keywords in them.


\subsection{Keyword Visualizaion of \method}

\begin{figure}
    \centering
    \includegraphics[width=1\linewidth]{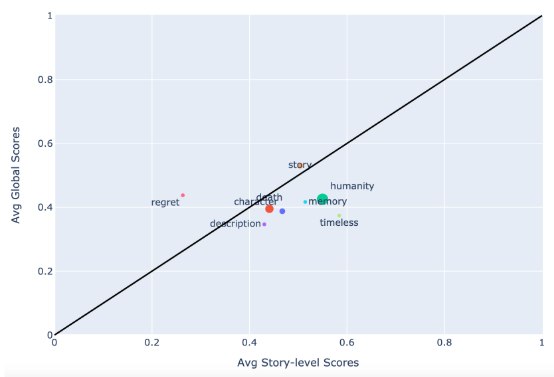}
    \caption{Suppose we want to understand the prediction rating of the \textit{heartwarming/touch} aspect for a stroy,
    we can visualize the generated keywords in all of the generated samples. The x-axis plots the average rating of the keyword for this story, and the y-axis plots the global rating of the keyword averaged across the training set. The size of the keyword proportional to its frequency in the generated keyword sequences.}
    \label{fig:method_visualization}
\end{figure}

A scorer without the help of a rationalizer could only provide a rating prediction for each aspect and users often want to know where the rating comes from. The keywords in \method allow user to visualize what causes the final rating prediction. For instance, \autoref{fig:method_visualization} illustrates that \textit{humanity} tends to be a negative keyword in the training data but being a positive keyword for the \textit{heartwarming} aspect of this story, so the depiction of the \textit{humanity} in this story increase its final \textit{touching} rating. 




\section{Related Work}

Due to the importance of automatic story evaluation, several types of approaches have been proposed. ROUGE~\citep{lin2004rouge}, BERTScore~\citep{zhang2019bertscore}, BARTScore~\citep{yuan2021bartscore}, and CTC~\citep{deng2021compression} compare the similarity between the generated text and the reference story. 
Although being effective in many other text generation tasks, higher similarity to the reference story is not necessarily a better story. Another type of evaluation method injects some noise into the human-written stories to create the low-quality stories and train a classifier to separate them.
Examples include UNION~\citep{guan2020union}, MANPLTS~\citep{ghazarian2021plot}, UNIEVAL~\citep{zhong2022towards}, and DELTAScore~\citep{xie2023deltascore}. 
Although these methods are good at discovering the incoherency from smaller language models, they cannot be used to evaluate a human-written story given a fine-grained aspect. 
Recently, researchers propose many general-purpose evaluation methods based on LLMs.
For example, GPTScore~\citep{fu2023gptscore} and G-Eval~\citep{liu2023g} directly prompt the LLM and several open-source models distill LLMs to reduce the evaluation cost~\citep{gao2024llm}. \citet{li2024leveraging,li2024llms} summarize these LLM-as-judge studies well.
In these papers, GPT-4 usually demonstrates the best correlation with human judgments.

Methodologically, our method is related to the LLM rationale generation and Minimum Bayes Risk (MBR) decoding~\citep{bertsch2023s}. 
Recent work in generating fluent free-text rationales has made use of two types of approaches - fine-tuning a small language model with gold human written rationales \citep{camburu2018esnli,narang2020wt5, wiegreffe-etal-2021-measuring} or zero-shot prompting LLMs to generate free-text rationales \citep{jung2022maieutic,wei2023chainofthought, kojima2023large, li2023making,lightman2023lets}.
Some approaches also leverage few-shot training approaches with a handful of gold rationales \citep{marasović2022fewshot, chen2023zara}. 
Our method could also be viewed as a special case of MBR, which generally refers to the methods that merge multiple generated candidate answers to improve the output quality. 
Other special cases of MRB include self-consistency prompting~\citep{wang2022self}, crowd sampling~\citep{suzgun2023follow}, complex CoT~\citep{fu2022complexity}, and output ensembling~\citep{martinez2023amrs}.

\section{Conclusion}

In this study, we look at a simple, yet efficient way to evaluate story-aspect pairs.
We propose \method that samples multiple generated keyword sequences before explanations, and using the generated keywords to score an aspect-story pair.
We posit that sampling helps us get diverse annotator ratings, and using keywords helps alleviate the objective mismatch between generating coherent explanations vs. usable explanations for downstream scoring. 
We show that that keywords not only improve the rating prediction performances, but also make the evaluation more interpretable and controllable. 


\section*{Limitations}
This work focuses on the fine-grain story evaluation task, which causes two limitations. First, we do not know if \method could also improve CoT in the other applications that involve subjective human judgements. Second, our choice of evaluation dataset is limited and it is hard to know if \method could bring similar improvements in other types of stories. 

In \Cref{tab:ablation}, we show that increasing the sizes of rationalizer could lead to better performance, but we do not have resources to fine-tune the LMs that are larger than 3b. Furthermore, most of our experiments in this work, while still relevant, are done before early 2024, so we did not evaluate the performance of large reasoning models such as o1 or o3. Nevertheless, reasoning models are expensive and not optimized for such subjective tasks, so \method should still be state-of-the-art method in fine-grained story evaluation, especially when we consider the inference cost.

Finally, there are some more complex LLM-as-judges approaches. For example, \citet{verga2024replacing} show that prompting multiple LLMs to discuss with each other improves the quality and reduces the cost of the evaluation task. However, we believe that the large performance gap between \method and the off-the-shelf LLMs in \Cref{tab:main_baseline_results} demonstrate the prompting LLMs without customizing/fine-tuning the LLMs is not very likely to achieve state-of-the-art results in subjective story evaluation tasks.





\section*{Ethical Statement and Broader Impact}
When dealing with ambiguity in evaluation tasks, one of the most common methods is to collect more fine-grained annotations~\citep{wu2024fine}. However, our work shows that some story evaluation tasks are so subjective that only collecting fine-grained annotations is not sufficient. 

The rising of the large reasoning models demonstrates the potential of LLMs given a high quality evaluation model. Nevertheless, no reliable reward model exists in more subjective tasks such as story evaluation. Our work could potentially provide some useful clues for solving the great challenge.

Finally, although customizing evaluation model is necessary in some applications, consistently targeting audience might intensify the problems of the filter bubbles~\citep{spohr2017fake}. For example, using \method to filter the story submissions could reduce the manually reviewing cost and make reviewing much more submissions possible, but it could also intensify the selection biases in the dataset that trains the evaluation model.

\section{Acknowledgments}
We thank Anjali Narayan-Chen, Alessandra Cervone and Shereen Oraby for discussions during this project. We also thank the USC INK Lab and Xiang Ren for feedback on the draft and work. Additionally, we thank anonymous reviewers and ACs for their feedback on improving this work.





\bibliography{custom}

\begin{thebibliography}{58}
\providecommand{\natexlab}[1]{#1}

\bibitem[{Bertsch et~al.(2023)Bertsch, Xie, Neubig, and Gormley}]{bertsch2023s}
Amanda Bertsch, Alex Xie, Graham Neubig, and Matthew~R Gormley. 2023.
\newblock It’s mbr all the way down: Modern generation techniques through the
  lens of minimum bayes risk.
\newblock In \emph{Proceedings of the Big Picture Workshop}, pages 108--122.

\bibitem[{Brown et~al.(2020)Brown, Mann, Ryder, Subbiah, Kaplan, Dhariwal,
  Neelakantan, Shyam, Sastry, Askell, Agarwal, Herbert-Voss, Krueger, Henighan,
  Child, Ramesh, Ziegler, Wu, Winter, Hesse, Chen, Sigler, Litwin, Gray, Chess,
  Clark, Berner, McCandlish, Radford, Sutskever, and
  Amodei}]{brown2020language}
Tom~B. Brown, Benjamin Mann, Nick Ryder, Melanie Subbiah, Jared Kaplan,
  Prafulla Dhariwal, Arvind Neelakantan, Pranav Shyam, Girish Sastry, Amanda
  Askell, Sandhini Agarwal, Ariel Herbert-Voss, Gretchen Krueger, Tom Henighan,
  Rewon Child, Aditya Ramesh, Daniel~M. Ziegler, Jeffrey Wu, Clemens Winter,
  Christopher Hesse, Mark Chen, Eric Sigler, Mateusz Litwin, Scott Gray,
  Benjamin Chess, Jack Clark, Christopher Berner, Sam McCandlish, Alec Radford,
  Ilya Sutskever, and Dario Amodei. 2020.
\newblock \href {https://arxiv.org/abs/2005.14165} {Language models are
  few-shot learners}.
\newblock \emph{Preprint}, arXiv:2005.14165.

\bibitem[{Camburu et~al.(2018)Camburu, Rockt\"{a}schel, Lukasiewicz, and
  Blunsom}]{camburu2018esnli}
Oana-Maria Camburu, Tim Rockt\"{a}schel, Thomas Lukasiewicz, and Phil Blunsom.
  2018.
\newblock \href
  {https://proceedings.neurips.cc/paper_files/paper/2018/file/4c7a167bb329bd92580a99ce422d6fa6-Paper.pdf}
  {e-snli: Natural language inference with natural language explanations}.
\newblock In \emph{Advances in Neural Information Processing Systems},
  volume~31. Curran Associates, Inc.

\bibitem[{Chen et~al.(2022)Chen, Vo, Takamura, Miyao, and
  Nakayama}]{chen-etal-2022-storyer}
Hong Chen, Duc Vo, Hiroya Takamura, Yusuke Miyao, and Hideki Nakayama. 2022.
\newblock \href {https://doi.org/10.18653/v1/2022.emnlp-main.114} {{S}tory{ER}:
  Automatic story evaluation via ranking, rating and reasoning}.
\newblock In \emph{Proceedings of the 2022 Conference on Empirical Methods in
  Natural Language Processing}, pages 1739--1753, Abu Dhabi, United Arab
  Emirates. Association for Computational Linguistics.

\bibitem[{Chen et~al.(2023)Chen, Yen, Huang, Wu, and Chen}]{chen2023zara}
Wei-Lin Chen, An-Zi Yen, Hen-Hsen Huang, Cheng-Kuang Wu, and Hsin-Hsi Chen.
  2023.
\newblock \href {https://arxiv.org/abs/2305.07355} {Zara: Improving few-shot
  self-rationalization for small language models}.
\newblock \emph{Preprint}, arXiv:2305.07355.

\bibitem[{Cicchetti(1994)}]{iccscore}
Domenic Cicchetti. 1994.
\newblock \href {https://doi.org/10.1037/1040-3590.6.4.284} {Guidelines,
  criteria, and rules of thumb for evaluating normed and standardized
  assessment instrument in psychology}.
\newblock \emph{Psychological Assessment}, 6:284--290.

\bibitem[{Deng et~al.(2021)Deng, Tan, Liu, Xing, and Hu}]{deng2021compression}
Mingkai Deng, Bowen Tan, Zhengzhong Liu, Eric Xing, and Zhiting Hu. 2021.
\newblock Compression, transduction, and creation: A unified framework for
  evaluating natural language generation.
\newblock In \emph{Proceedings of the 2021 Conference on Empirical Methods in
  Natural Language Processing}, pages 7580--7605.

\bibitem[{Doostmohammadi et~al.(2024)Doostmohammadi, Holmstr{\"o}m, and
  Kuhlmann}]{doostmohammadi2024reliable}
Ehsan Doostmohammadi, Oskar Holmstr{\"o}m, and Marco Kuhlmann. 2024.
\newblock How reliable are automatic evaluation methods for instruction-tuned
  llms?
\newblock \emph{arXiv preprint arXiv:2402.10770}.

\bibitem[{Fan et~al.(2018)Fan, Lewis, and Dauphin}]{fan-etal-2018-hierarchical}
Angela Fan, Mike Lewis, and Yann Dauphin. 2018.
\newblock \href {https://doi.org/10.18653/v1/P18-1082} {Hierarchical neural
  story generation}.
\newblock In \emph{Proceedings of the 56th Annual Meeting of the Association
  for Computational Linguistics (Volume 1: Long Papers)}, pages 889--898,
  Melbourne, Australia. Association for Computational Linguistics.

\bibitem[{Florian et~al.(2024)Florian, Alexandre, Benjamin, Yann, and
  Alexandre}]{florian2024exploring}
Le~Bronnec Florian, Verine Alexandre, Negrevergne Benjamin, Chevaleyre Yann,
  and Allauzen Alexandre. 2024.
\newblock Exploring precision and recall to assess the quality and diversity of
  llms.
\newblock \emph{arXiv preprint arXiv:2402.10693}.

\bibitem[{Frank et~al.(1999)Frank, Paynter, Witten, Gutwin, and
  Nevill-Manning}]{frank1999domain}
Eibe Frank, Gordon~W Paynter, Ian~H Witten, Carl Gutwin, and Craig~G
  Nevill-Manning. 1999.
\newblock Domain-specific keyphrase extraction. jcai’99: Proceedings of the
  sixteenth international joint conference on artificial intelligence, 668-673.

\bibitem[{Fu et~al.(2023)Fu, Ng, Jiang, and Liu}]{fu2023gptscore}
Jinlan Fu, See-Kiong Ng, Zhengbao Jiang, and Pengfei Liu. 2023.
\newblock Gptscore: Evaluate as you desire.
\newblock \emph{arXiv preprint arXiv:2302.04166}.

\bibitem[{Fu et~al.(2022)Fu, Peng, Sabharwal, Clark, and
  Khot}]{fu2022complexity}
Yao Fu, Hao Peng, Ashish Sabharwal, Peter Clark, and Tushar Khot. 2022.
\newblock Complexity-based prompting for multi-step reasoning.
\newblock In \emph{The Eleventh International Conference on Learning
  Representations}.

\bibitem[{Gao et~al.(2024)Gao, Hu, Ruan, Pu, and Wan}]{gao2024llm}
Mingqi Gao, Xinyu Hu, Jie Ruan, Xiao Pu, and Xiaojun Wan. 2024.
\newblock Llm-based nlg evaluation: Current status and challenges.
\newblock \emph{arXiv preprint arXiv:2402.01383}.

\bibitem[{Ghazarian et~al.(2021)Ghazarian, Liu, Akash, Weischedel, Galstyan,
  and Peng}]{ghazarian2021plot}
Sarik Ghazarian, Zixi Liu, SM~Akash, Ralph Weischedel, Aram Galstyan, and
  Nanyun Peng. 2021.
\newblock Plot-guided adversarial example construction for evaluating
  open-domain story generation.
\newblock In \emph{Proceedings of the 2021 Conference of the North American
  Chapter of the Association for Computational Linguistics: Human Language
  Technologies}, pages 4334--4344.

\bibitem[{Gmyrek et~al.(2024)Gmyrek, Lutz, and
  Newlands}]{gmyrek2024technological}
Pawel Gmyrek, Christoph Lutz, and Gemma Newlands. 2024.
\newblock A technological construction of society: Comparing gpt-4 and human
  respondents for occupational evaluation in the uk.
\newblock \emph{Available at SSRN 4700366}.

\bibitem[{Guan and Huang(2020)}]{guan2020union}
Jian Guan and Minlie Huang. 2020.
\newblock Union: An unreferenced metric for evaluating open-ended story
  generation.
\newblock In \emph{Proceedings of the 2020 Conference on Empirical Methods in
  Natural Language Processing (EMNLP)}, pages 9157--9166.

\bibitem[{G{\"u}lich and Quasthoff(1986)}]{gulich1986story}
Elisabeth G{\"u}lich and Uta~M Quasthoff. 1986.
\newblock Story-telling in conversation: cognitive and interactive aspects.
\newblock \emph{Poetics}, 15(1-2):217--241.

\bibitem[{He et~al.(2021)He, Gao, and Chen}]{he2021debertav3}
Pengcheng He, Jianfeng Gao, and Weizhu Chen. 2021.
\newblock \href {https://arxiv.org/abs/2111.09543} {Debertav3: Improving
  deberta using electra-style pre-training with gradient-disentangled embedding
  sharing}.
\newblock \emph{Preprint}, arXiv:2111.09543.

\bibitem[{Jain et~al.(2020)Jain, Wiegreffe, Pinter, and
  Wallace}]{jain-etal-2020-learning}
Sarthak Jain, Sarah Wiegreffe, Yuval Pinter, and Byron~C. Wallace. 2020.
\newblock \href {https://doi.org/10.18653/v1/2020.acl-main.409} {{L}earning to
  faithfully rationalize by construction}.
\newblock In \emph{Proceedings of the 58th Annual Meeting of the Association
  for Computational Linguistics}, pages 4459--4473, Online. Association for
  Computational Linguistics.

\bibitem[{Jia et~al.(2020)Jia, Meng, Zhao, and
  Chang}]{jia-etal-2020-mitigating}
Shengyu Jia, Tao Meng, Jieyu Zhao, and Kai-Wei Chang. 2020.
\newblock \href {https://doi.org/10.18653/v1/2020.acl-main.264} {Mitigating
  gender bias amplification in distribution by posterior regularization}.
\newblock In \emph{Proceedings of the 58th Annual Meeting of the Association
  for Computational Linguistics}, pages 2936--2942, Online. Association for
  Computational Linguistics.

\bibitem[{Jiang et~al.(2023)Jiang, Sablayrolles, Mensch, Bamford, Chaplot,
  de~las Casas, Bressand, Lengyel, Lample, Saulnier, Lavaud, Lachaux, Stock,
  Scao, Lavril, Wang, Lacroix, and Sayed}]{jiang2023mistral}
Albert~Q. Jiang, Alexandre Sablayrolles, Arthur Mensch, Chris Bamford,
  Devendra~Singh Chaplot, Diego de~las Casas, Florian Bressand, Gianna Lengyel,
  Guillaume Lample, Lucile Saulnier, Lélio~Renard Lavaud, Marie-Anne Lachaux,
  Pierre Stock, Teven~Le Scao, Thibaut Lavril, Thomas Wang, Timothée Lacroix,
  and William~El Sayed. 2023.
\newblock \href {https://arxiv.org/abs/2310.06825} {Mistral 7b}.
\newblock \emph{Preprint}, arXiv:2310.06825.

\bibitem[{Jung et~al.(2022)Jung, Qin, Welleck, Brahman, Bhagavatula, Bras, and
  Choi}]{jung2022maieutic}
Jaehun Jung, Lianhui Qin, Sean Welleck, Faeze Brahman, Chandra Bhagavatula,
  Ronan~Le Bras, and Yejin Choi. 2022.
\newblock Maieutic prompting: Logically consistent reasoning with recursive
  explanations.
\newblock \emph{arXiv preprint arXiv:2205.11822}.

\bibitem[{Kim et~al.(2023)Kim, Joo, Kim, Jang, Ye, Shin, and
  Seo}]{kim-etal-2023-cot}
Seungone Kim, Se~Joo, Doyoung Kim, Joel Jang, Seonghyeon Ye, Jamin Shin, and
  Minjoon Seo. 2023.
\newblock \href {https://doi.org/10.18653/v1/2023.emnlp-main.782} {The {C}o{T}
  collection: Improving zero-shot and few-shot learning of language models via
  chain-of-thought fine-tuning}.
\newblock In \emph{Proceedings of the 2023 Conference on Empirical Methods in
  Natural Language Processing}, pages 12685--12708, Singapore. Association for
  Computational Linguistics.

\bibitem[{Kirk et~al.(2024)Kirk, Mediratta, Nalmpantis, Luketina, Hambro,
  Grefenstette, and Raileanu}]{kirk2024understanding}
Robert Kirk, Ishita Mediratta, Christoforos Nalmpantis, Jelena Luketina, Eric
  Hambro, Edward Grefenstette, and Roberta Raileanu. 2024.
\newblock \href {https://openreview.net/forum?id=PXD3FAVHJT} {Understanding the
  effects of {RLHF} on {LLM} generalisation and diversity}.
\newblock In \emph{The Twelfth International Conference on Learning
  Representations}.

\bibitem[{Kojima et~al.(2023)Kojima, Gu, Reid, Matsuo, and
  Iwasawa}]{kojima2023large}
Takeshi Kojima, Shixiang~Shane Gu, Machel Reid, Yutaka Matsuo, and Yusuke
  Iwasawa. 2023.
\newblock \href {https://arxiv.org/abs/2205.11916} {Large language models are
  zero-shot reasoners}.
\newblock \emph{Preprint}, arXiv:2205.11916.

\bibitem[{Li et~al.(2024{\natexlab{a}})Li, Dong, Chen, Su, Zhou, Ai, Ye, and
  Liu}]{li2024llms}
Haitao Li, Qian Dong, Junjie Chen, Huixue Su, Yujia Zhou, Qingyao Ai, Ziyi Ye,
  and Yiqun Liu. 2024{\natexlab{a}}.
\newblock Llms-as-judges: a comprehensive survey on llm-based evaluation
  methods.
\newblock \emph{arXiv preprint arXiv:2412.05579}.

\bibitem[{Li et~al.(2023)Li, Lin, Zhang, Fu, Chen, Lou, and
  Chen}]{li2023making}
Yifei Li, Zeqi Lin, Shizhuo Zhang, Qiang Fu, Bei Chen, Jian-Guang Lou, and
  Weizhu Chen. 2023.
\newblock Making language models better reasoners with step-aware verifier.
\newblock In \emph{Proceedings of the 61st Annual Meeting of the Association
  for Computational Linguistics (Volume 1: Long Papers)}, pages 5315--5333.

\bibitem[{Li et~al.(2024{\natexlab{b}})Li, Xu, Shen, Xu, Gu, and
  Tao}]{li2024leveraging}
Zhen Li, Xiaohan Xu, Tao Shen, Can Xu, Jia-Chen Gu, and Chongyang Tao.
  2024{\natexlab{b}}.
\newblock Leveraging large language models for nlg evaluation: A survey.
\newblock \emph{arXiv preprint arXiv:2401.07103}.

\bibitem[{Lightman et~al.(2023)Lightman, Kosaraju, Burda, Edwards, Baker, Lee,
  Leike, Schulman, Sutskever, and Cobbe}]{lightman2023lets}
Hunter Lightman, Vineet Kosaraju, Yura Burda, Harri Edwards, Bowen Baker, Teddy
  Lee, Jan Leike, John Schulman, Ilya Sutskever, and Karl Cobbe. 2023.
\newblock \href {https://arxiv.org/abs/2305.20050} {Let's verify step by step}.
\newblock \emph{Preprint}, arXiv:2305.20050.

\bibitem[{Lin(2004)}]{lin2004rouge}
Chin-Yew Lin. 2004.
\newblock \href {https://aclanthology.org/W04-1013} {{ROUGE}: A package for
  automatic evaluation of summaries}.
\newblock In \emph{Text Summarization Branches Out}, pages 74--81, Barcelona,
  Spain. Association for Computational Linguistics.

\bibitem[{Liu et~al.(2023)Liu, Iter, Xu, Wang, Xu, and Zhu}]{liu2023g}
Yang Liu, Dan Iter, Yichong Xu, Shuohang Wang, Ruochen Xu, and Chenguang Zhu.
  2023.
\newblock G-eval: Nlg evaluation using gpt-4 with better human alignment.
\newblock In \emph{The 2023 Conference on Empirical Methods in Natural Language
  Processing}.

\bibitem[{Marasović et~al.(2022)Marasović, Beltagy, Downey, and
  Peters}]{marasović2022fewshot}
Ana Marasović, Iz~Beltagy, Doug Downey, and Matthew~E. Peters. 2022.
\newblock \href {https://arxiv.org/abs/2111.08284} {Few-shot
  self-rationalization with natural language prompts}.
\newblock \emph{Preprint}, arXiv:2111.08284.

\bibitem[{Martinez~Lorenzo et~al.(2023)Martinez~Lorenzo, Huguet~Cabot, Navigli
  et~al.}]{martinez2023amrs}
Abelardo~Carlos Martinez~Lorenzo, Pere~Llu{\'\i}s Huguet~Cabot, Roberto
  Navigli, et~al. 2023.
\newblock Amrs assemble! learning to ensemble with autoregressive models for
  amr parsing.
\newblock In \emph{Proceedings of the 61st Annual Meeting of the Association
  for Computational Linguistics (Volume 2: Short Papers)}, pages 1595--1605.

\bibitem[{Mihalcea and Tarau(2004)}]{mihalcea2004textrank}
Rada Mihalcea and Paul Tarau. 2004.
\newblock Textrank: Bringing order into text.
\newblock In \emph{Proceedings of the 2004 conference on empirical methods in
  natural language processing}, pages 404--411.

\bibitem[{Narang et~al.(2020)Narang, Raffel, Lee, Roberts, Fiedel, and
  Malkan}]{narang2020wt5}
Sharan Narang, Colin Raffel, Katherine Lee, Adam Roberts, Noah Fiedel, and
  Karishma Malkan. 2020.
\newblock Wt5?! training text-to-text models to explain their predictions.
\newblock \emph{arXiv preprint arXiv:2004.14546}.

\bibitem[{OpenAI et~al.(2024)OpenAI, Achiam, Adler, Agarwal, Ahmad, Akkaya,
  Aleman, Almeida, Altenschmidt, Altman, Anadkat, Avila, Babuschkin, Balaji,
  Balcom, Baltescu, Bao, Bavarian, Belgum, Bello, Berdine, Bernadett-Shapiro,
  Berner, Bogdonoff, Boiko, Boyd, Brakman, Brockman, Brooks, Brundage, Button,
  Cai, Campbell, Cann, Carey, Carlson, Carmichael, Chan, Chang, Chantzis, Chen,
  Chen, Chen, Chen, Chen, Chess, Cho, Chu, Chung, Cummings, Currier, Dai,
  Decareaux, Degry, Deutsch, Deville, Dhar, Dohan, Dowling, Dunning, Ecoffet,
  Eleti, Eloundou, Farhi, Fedus, Felix, Fishman, Forte, Fulford, Gao, Georges,
  Gibson, Goel, Gogineni, Goh, Gontijo-Lopes, Gordon, Grafstein, Gray, Greene,
  Gross, Gu, Guo, Hallacy, Han, Harris, He, Heaton, Heidecke, Hesse, Hickey,
  Hickey, Hoeschele, Houghton, Hsu, Hu, Hu, Huizinga, Jain, Jain, Jang, Jiang,
  Jiang, Jin, Jin, Jomoto, Jonn, Jun, Kaftan, Łukasz Kaiser, Kamali,
  Kanitscheider, Keskar, Khan, Kilpatrick, Kim, Kim, Kim, Kirchner, Kiros,
  Knight, Kokotajlo, Łukasz Kondraciuk, Kondrich, Konstantinidis, Kosic,
  Krueger, Kuo, Lampe, Lan, Lee, Leike, Leung, Levy, Li, Lim, Lin, Lin, Litwin,
  Lopez, Lowe, Lue, Makanju, Malfacini, Manning, Markov, Markovski, Martin,
  Mayer, Mayne, McGrew, McKinney, McLeavey, McMillan, McNeil, Medina, Mehta,
  Menick, Metz, Mishchenko, Mishkin, Monaco, Morikawa, Mossing, Mu, Murati,
  Murk, Mély, Nair, Nakano, Nayak, Neelakantan, Ngo, Noh, Ouyang, O'Keefe,
  Pachocki, Paino, Palermo, Pantuliano, Parascandolo, Parish, Parparita,
  Passos, Pavlov, Peng, Perelman, de~Avila Belbute~Peres, Petrov,
  de~Oliveira~Pinto, Michael, Pokorny, Pokrass, Pong, Powell, Power, Power,
  Proehl, Puri, Radford, Rae, Ramesh, Raymond, Real, Rimbach, Ross, Rotsted,
  Roussez, Ryder, Saltarelli, Sanders, Santurkar, Sastry, Schmidt, Schnurr,
  Schulman, Selsam, Sheppard, Sherbakov, Shieh, Shoker, Shyam, Sidor, Sigler,
  Simens, Sitkin, Slama, Sohl, Sokolowsky, Song, Staudacher, Such, Summers,
  Sutskever, Tang, Tezak, Thompson, Tillet, Tootoonchian, Tseng, Tuggle,
  Turley, Tworek, Uribe, Vallone, Vijayvergiya, Voss, Wainwright, Wang, Wang,
  Wang, Ward, Wei, Weinmann, Welihinda, Welinder, Weng, Weng, Wiethoff,
  Willner, Winter, Wolrich, Wong, Workman, Wu, Wu, Wu, Xiao, Xu, Yoo, Yu, Yuan,
  Zaremba, Zellers, Zhang, Zhang, Zhao, Zheng, Zhuang, Zhuk, and
  Zoph}]{openai2024gpt4}
OpenAI, Josh Achiam, Steven Adler, Sandhini Agarwal, Lama Ahmad, Ilge Akkaya,
  Florencia~Leoni Aleman, Diogo Almeida, Janko Altenschmidt, Sam Altman,
  Shyamal Anadkat, Red Avila, Igor Babuschkin, Suchir Balaji, Valerie Balcom,
  Paul Baltescu, Haiming Bao, Mohammad Bavarian, Jeff Belgum, Irwan Bello, Jake
  Berdine, Gabriel Bernadett-Shapiro, Christopher Berner, Lenny Bogdonoff, Oleg
  Boiko, Madelaine Boyd, Anna-Luisa Brakman, Greg Brockman, Tim Brooks, Miles
  Brundage, Kevin Button, Trevor Cai, Rosie Campbell, Andrew Cann, Brittany
  Carey, Chelsea Carlson, Rory Carmichael, Brooke Chan, Che Chang, Fotis
  Chantzis, Derek Chen, Sully Chen, Ruby Chen, Jason Chen, Mark Chen, Ben
  Chess, Chester Cho, Casey Chu, Hyung~Won Chung, Dave Cummings, Jeremiah
  Currier, Yunxing Dai, Cory Decareaux, Thomas Degry, Noah Deutsch, Damien
  Deville, Arka Dhar, David Dohan, Steve Dowling, Sheila Dunning, Adrien
  Ecoffet, Atty Eleti, Tyna Eloundou, David Farhi, Liam Fedus, Niko Felix,
  Simón~Posada Fishman, Juston Forte, Isabella Fulford, Leo Gao, Elie Georges,
  Christian Gibson, Vik Goel, Tarun Gogineni, Gabriel Goh, Rapha Gontijo-Lopes,
  Jonathan Gordon, Morgan Grafstein, Scott Gray, Ryan Greene, Joshua Gross,
  Shixiang~Shane Gu, Yufei Guo, Chris Hallacy, Jesse Han, Jeff Harris, Yuchen
  He, Mike Heaton, Johannes Heidecke, Chris Hesse, Alan Hickey, Wade Hickey,
  Peter Hoeschele, Brandon Houghton, Kenny Hsu, Shengli Hu, Xin Hu, Joost
  Huizinga, Shantanu Jain, Shawn Jain, Joanne Jang, Angela Jiang, Roger Jiang,
  Haozhun Jin, Denny Jin, Shino Jomoto, Billie Jonn, Heewoo Jun, Tomer Kaftan,
  Łukasz Kaiser, Ali Kamali, Ingmar Kanitscheider, Nitish~Shirish Keskar,
  Tabarak Khan, Logan Kilpatrick, Jong~Wook Kim, Christina Kim, Yongjik Kim,
  Jan~Hendrik Kirchner, Jamie Kiros, Matt Knight, Daniel Kokotajlo, Łukasz
  Kondraciuk, Andrew Kondrich, Aris Konstantinidis, Kyle Kosic, Gretchen
  Krueger, Vishal Kuo, Michael Lampe, Ikai Lan, Teddy Lee, Jan Leike, Jade
  Leung, Daniel Levy, Chak~Ming Li, Rachel Lim, Molly Lin, Stephanie Lin,
  Mateusz Litwin, Theresa Lopez, Ryan Lowe, Patricia Lue, Anna Makanju, Kim
  Malfacini, Sam Manning, Todor Markov, Yaniv Markovski, Bianca Martin, Katie
  Mayer, Andrew Mayne, Bob McGrew, Scott~Mayer McKinney, Christine McLeavey,
  Paul McMillan, Jake McNeil, David Medina, Aalok Mehta, Jacob Menick, Luke
  Metz, Andrey Mishchenko, Pamela Mishkin, Vinnie Monaco, Evan Morikawa, Daniel
  Mossing, Tong Mu, Mira Murati, Oleg Murk, David Mély, Ashvin Nair, Reiichiro
  Nakano, Rajeev Nayak, Arvind Neelakantan, Richard Ngo, Hyeonwoo Noh, Long
  Ouyang, Cullen O'Keefe, Jakub Pachocki, Alex Paino, Joe Palermo, Ashley
  Pantuliano, Giambattista Parascandolo, Joel Parish, Emy Parparita, Alex
  Passos, Mikhail Pavlov, Andrew Peng, Adam Perelman, Filipe de~Avila
  Belbute~Peres, Michael Petrov, Henrique~Ponde de~Oliveira~Pinto, Michael,
  Pokorny, Michelle Pokrass, Vitchyr~H. Pong, Tolly Powell, Alethea Power,
  Boris Power, Elizabeth Proehl, Raul Puri, Alec Radford, Jack Rae, Aditya
  Ramesh, Cameron Raymond, Francis Real, Kendra Rimbach, Carl Ross, Bob
  Rotsted, Henri Roussez, Nick Ryder, Mario Saltarelli, Ted Sanders, Shibani
  Santurkar, Girish Sastry, Heather Schmidt, David Schnurr, John Schulman,
  Daniel Selsam, Kyla Sheppard, Toki Sherbakov, Jessica Shieh, Sarah Shoker,
  Pranav Shyam, Szymon Sidor, Eric Sigler, Maddie Simens, Jordan Sitkin,
  Katarina Slama, Ian Sohl, Benjamin Sokolowsky, Yang Song, Natalie Staudacher,
  Felipe~Petroski Such, Natalie Summers, Ilya Sutskever, Jie Tang, Nikolas
  Tezak, Madeleine~B. Thompson, Phil Tillet, Amin Tootoonchian, Elizabeth
  Tseng, Preston Tuggle, Nick Turley, Jerry Tworek, Juan Felipe~Cerón Uribe,
  Andrea Vallone, Arun Vijayvergiya, Chelsea Voss, Carroll Wainwright,
  Justin~Jay Wang, Alvin Wang, Ben Wang, Jonathan Ward, Jason Wei, CJ~Weinmann,
  Akila Welihinda, Peter Welinder, Jiayi Weng, Lilian Weng, Matt Wiethoff, Dave
  Willner, Clemens Winter, Samuel Wolrich, Hannah Wong, Lauren Workman, Sherwin
  Wu, Jeff Wu, Michael Wu, Kai Xiao, Tao Xu, Sarah Yoo, Kevin Yu, Qiming Yuan,
  Wojciech Zaremba, Rowan Zellers, Chong Zhang, Marvin Zhang, Shengjia Zhao,
  Tianhao Zheng, Juntang Zhuang, William Zhuk, and Barret Zoph. 2024.
\newblock \href {https://arxiv.org/abs/2303.08774} {Gpt-4 technical report}.
\newblock \emph{Preprint}, arXiv:2303.08774.

\bibitem[{Ouyang et~al.(2022)Ouyang, Wu, Jiang, Almeida, Wainwright, Mishkin,
  Zhang, Agarwal, Slama, Ray, Schulman, Hilton, Kelton, Miller, Simens, Askell,
  Welinder, Christiano, Leike, and Lowe}]{ouyang2022training}
Long Ouyang, Jeff Wu, Xu~Jiang, Diogo Almeida, Carroll~L. Wainwright, Pamela
  Mishkin, Chong Zhang, Sandhini Agarwal, Katarina Slama, Alex Ray, John
  Schulman, Jacob Hilton, Fraser Kelton, Luke Miller, Maddie Simens, Amanda
  Askell, Peter Welinder, Paul Christiano, Jan Leike, and Ryan Lowe. 2022.
\newblock \href {https://arxiv.org/abs/2203.02155} {Training language models to
  follow instructions with human feedback}.
\newblock \emph{Preprint}, arXiv:2203.02155.

\bibitem[{Raffel et~al.(2023)Raffel, Shazeer, Roberts, Lee, Narang, Matena,
  Zhou, Li, and Liu}]{raffel2023exploring}
Colin Raffel, Noam Shazeer, Adam Roberts, Katherine Lee, Sharan Narang, Michael
  Matena, Yanqi Zhou, Wei Li, and Peter~J. Liu. 2023.
\newblock \href {https://arxiv.org/abs/1910.10683} {Exploring the limits of
  transfer learning with a unified text-to-text transformer}.
\newblock \emph{Preprint}, arXiv:1910.10683.

\bibitem[{Ramamurthy et~al.(2022)Ramamurthy, Ammanabrolu, Brantley, Hessel,
  Sifa, Bauckhage, Hajishirzi, and Choi}]{Ramamurthy2022IsRL}
Rajkumar Ramamurthy, Prithviraj Ammanabrolu, Kiant{\'e} Brantley, Jack Hessel,
  Rafet Sifa, Christian Bauckhage, Hannaneh Hajishirzi, and Yejin Choi. 2022.
\newblock \href {https://arxiv.org/abs/2210.01241} {Is reinforcement learning
  (not) for natural language processing?: Benchmarks, baselines, and building
  blocks for natural language policy optimization}.

\bibitem[{Rose et~al.(2010)Rose, Engel, Cramer, and Cowley}]{rose2010automatic}
Stuart Rose, Dave Engel, Nick Cramer, and Wendy Cowley. 2010.
\newblock Automatic keyword extraction from individual documents.
\newblock \emph{Text mining: applications and theory}, pages 1--20.

\bibitem[{Sap et~al.(2021)Sap, Swayamdipta, Vianna, Zhou, Choi, and
  Smith}]{sap2021annotators}
Maarten Sap, Swabha Swayamdipta, Laura Vianna, Xuhui Zhou, Yejin Choi, and
  Noah~A Smith. 2021.
\newblock Annotators with attitudes: How annotator beliefs and identities bias
  toxic language detection.
\newblock \emph{arXiv preprint arXiv:2111.07997}.

\bibitem[{Schulman et~al.(2017)Schulman, Wolski, Dhariwal, Radford, and
  Klimov}]{schulman2017proximal}
John Schulman, Filip Wolski, Prafulla Dhariwal, Alec Radford, and Oleg Klimov.
  2017.
\newblock \href {https://arxiv.org/abs/1707.06347} {Proximal policy
  optimization algorithms}.
\newblock \emph{Preprint}, arXiv:1707.06347.

\bibitem[{Spohr(2017)}]{spohr2017fake}
Dominic Spohr. 2017.
\newblock Fake news and ideological polarization: Filter bubbles and selective
  exposure on social media.
\newblock \emph{Business information review}, 34(3):150--160.

\bibitem[{Suzgun et~al.(2023)Suzgun, Melas-Kyriazi, and
  Jurafsky}]{suzgun2023follow}
Mirac Suzgun, Luke Melas-Kyriazi, and Dan Jurafsky. 2023.
\newblock Follow the wisdom of the crowd: Effective text generation via minimum
  bayes risk decoding.
\newblock In \emph{Findings of the Association for Computational Linguistics:
  ACL 2023}, pages 4265--4293.

\bibitem[{Touvron et~al.(2023)Touvron, Martin, Stone, Albert, Almahairi,
  Babaei, Bashlykov, Batra, Bhargava, Bhosale, Bikel, Blecher, Ferrer, Chen,
  Cucurull, Esiobu, Fernandes, Fu, Fu, Fuller, Gao, Goswami, Goyal, Hartshorn,
  Hosseini, Hou, Inan, Kardas, Kerkez, Khabsa, Kloumann, Korenev, Koura,
  Lachaux, Lavril, Lee, Liskovich, Lu, Mao, Martinet, Mihaylov, Mishra,
  Molybog, Nie, Poulton, Reizenstein, Rungta, Saladi, Schelten, Silva, Smith,
  Subramanian, Tan, Tang, Taylor, Williams, Kuan, Xu, Yan, Zarov, Zhang, Fan,
  Kambadur, Narang, Rodriguez, Stojnic, Edunov, and Scialom}]{touvron2023llama}
Hugo Touvron, Louis Martin, Kevin Stone, Peter Albert, Amjad Almahairi, Yasmine
  Babaei, Nikolay Bashlykov, Soumya Batra, Prajjwal Bhargava, Shruti Bhosale,
  Dan Bikel, Lukas Blecher, Cristian~Canton Ferrer, Moya Chen, Guillem
  Cucurull, David Esiobu, Jude Fernandes, Jeremy Fu, Wenyin Fu, Brian Fuller,
  Cynthia Gao, Vedanuj Goswami, Naman Goyal, Anthony Hartshorn, Saghar
  Hosseini, Rui Hou, Hakan Inan, Marcin Kardas, Viktor Kerkez, Madian Khabsa,
  Isabel Kloumann, Artem Korenev, Punit~Singh Koura, Marie-Anne Lachaux,
  Thibaut Lavril, Jenya Lee, Diana Liskovich, Yinghai Lu, Yuning Mao, Xavier
  Martinet, Todor Mihaylov, Pushkar Mishra, Igor Molybog, Yixin Nie, Andrew
  Poulton, Jeremy Reizenstein, Rashi Rungta, Kalyan Saladi, Alan Schelten, Ruan
  Silva, Eric~Michael Smith, Ranjan Subramanian, Xiaoqing~Ellen Tan, Binh Tang,
  Ross Taylor, Adina Williams, Jian~Xiang Kuan, Puxin Xu, Zheng Yan, Iliyan
  Zarov, Yuchen Zhang, Angela Fan, Melanie Kambadur, Sharan Narang, Aurelien
  Rodriguez, Robert Stojnic, Sergey Edunov, and Thomas Scialom. 2023.
\newblock \href {https://arxiv.org/abs/2307.09288} {Llama 2: Open foundation
  and fine-tuned chat models}.
\newblock \emph{Preprint}, arXiv:2307.09288.

\bibitem[{Vedantam et~al.(2015)Vedantam, Zitnick, and
  Parikh}]{vedantam2015cider}
Ramakrishna Vedantam, C.~Lawrence Zitnick, and Devi Parikh. 2015.
\newblock \href {https://arxiv.org/abs/1411.5726} {Cider: Consensus-based image
  description evaluation}.
\newblock \emph{Preprint}, arXiv:1411.5726.

\bibitem[{Verga et~al.(2024)Verga, Hofstatter, Althammer, Su, Piktus,
  Arkhangorodsky, Xu, White, and Lewis}]{verga2024replacing}
Pat Verga, Sebastian Hofstatter, Sophia Althammer, Yixuan Su, Aleksandra
  Piktus, Arkady Arkhangorodsky, Minjie Xu, Naomi White, and Patrick Lewis.
  2024.
\newblock Replacing judges with juries: Evaluating llm generations with a panel
  of diverse models.
\newblock \emph{arXiv preprint arXiv:2404.18796}.

\bibitem[{Wang et~al.(2023{\natexlab{a}})Wang, Yang, Zhu, Yang, Cohen, Li, and
  Tian}]{wang2023learning}
Danqing Wang, Kevin Yang, Hanlin Zhu, Xiaomeng Yang, Andrew Cohen, Lei Li, and
  Yuandong Tian. 2023{\natexlab{a}}.
\newblock Learning personalized story evaluation.
\newblock \emph{arXiv preprint arXiv:2310.03304}.

\bibitem[{Wang et~al.(2023{\natexlab{b}})Wang, Wang, Li, Gao, Yin, and
  Ren}]{wang-etal-2023-scott}
Peifeng Wang, Zhengyang Wang, Zheng Li, Yifan Gao, Bing Yin, and Xiang Ren.
  2023{\natexlab{b}}.
\newblock \href {https://doi.org/10.18653/v1/2023.acl-long.304} {{SCOTT}:
  Self-consistent chain-of-thought distillation}.
\newblock In \emph{Proceedings of the 61st Annual Meeting of the Association
  for Computational Linguistics (Volume 1: Long Papers)}, pages 5546--5558,
  Toronto, Canada. Association for Computational Linguistics.

\bibitem[{Wang et~al.(2022)Wang, Wei, Schuurmans, Le, Chi, Narang, Chowdhery,
  and Zhou}]{wang2022self}
Xuezhi Wang, Jason Wei, Dale Schuurmans, Quoc~V Le, Ed~H Chi, Sharan Narang,
  Aakanksha Chowdhery, and Denny Zhou. 2022.
\newblock Self-consistency improves chain of thought reasoning in language
  models.
\newblock In \emph{The Eleventh International Conference on Learning
  Representations}.

\bibitem[{Wei et~al.(2023)Wei, Wang, Schuurmans, Bosma, Ichter, Xia, Chi, Le,
  and Zhou}]{wei2023chainofthought}
Jason Wei, Xuezhi Wang, Dale Schuurmans, Maarten Bosma, Brian Ichter, Fei Xia,
  Ed~Chi, Quoc Le, and Denny Zhou. 2023.
\newblock \href {https://arxiv.org/abs/2201.11903} {Chain-of-thought prompting
  elicits reasoning in large language models}.
\newblock \emph{Preprint}, arXiv:2201.11903.

\bibitem[{Wiegreffe et~al.(2021)Wiegreffe, Marasovi{\'c}, and
  Smith}]{wiegreffe-etal-2021-measuring}
Sarah Wiegreffe, Ana Marasovi{\'c}, and Noah~A. Smith. 2021.
\newblock \href {https://doi.org/10.18653/v1/2021.emnlp-main.804} {{M}easuring
  association between labels and free-text rationales}.
\newblock In \emph{Proceedings of the 2021 Conference on Empirical Methods in
  Natural Language Processing}, pages 10266--10284, Online and Punta Cana,
  Dominican Republic. Association for Computational Linguistics.

\bibitem[{Wu et~al.(2024)Wu, Hu, Shi, Dziri, Suhr, Ammanabrolu, Smith,
  Ostendorf, and Hajishirzi}]{wu2024fine}
Zeqiu Wu, Yushi Hu, Weijia Shi, Nouha Dziri, Alane Suhr, Prithviraj
  Ammanabrolu, Noah~A Smith, Mari Ostendorf, and Hannaneh Hajishirzi. 2024.
\newblock Fine-grained human feedback gives better rewards for language model
  training.
\newblock \emph{Advances in Neural Information Processing Systems}, 36.

\bibitem[{Xie et~al.(2023)Xie, Li, Cohn, and Lau}]{xie2023deltascore}
Zhuohan Xie, Miao Li, Trevor Cohn, and Jey~Han Lau. 2023.
\newblock Deltascore: Evaluating story generation with differentiating
  perturbations.
\newblock \emph{arXiv preprint arXiv:2303.08991}.

\bibitem[{Yuan et~al.(2021)Yuan, Neubig, and Liu}]{yuan2021bartscore}
Weizhe Yuan, Graham Neubig, and Pengfei Liu. 2021.
\newblock Bartscore: Evaluating generated text as text generation.
\newblock \emph{Advances in Neural Information Processing Systems},
  34:27263--27277.

\bibitem[{Zhang et~al.(2019)Zhang, Kishore, Wu, Weinberger, and
  Artzi}]{zhang2019bertscore}
Tianyi Zhang, Varsha Kishore, Felix Wu, Kilian~Q Weinberger, and Yoav Artzi.
  2019.
\newblock Bertscore: Evaluating text generation with bert.
\newblock In \emph{International Conference on Learning Representations}.

\bibitem[{Zhong et~al.(2022)Zhong, Liu, Yin, Mao, Jiao, Liu, Zhu, Ji, and
  Han}]{zhong2022towards}
Ming Zhong, Yang Liu, Da~Yin, Yuning Mao, Yizhu Jiao, Pengfei Liu, Chenguang
  Zhu, Heng Ji, and Jiawei Han. 2022.
\newblock Towards a unified multi-dimensional evaluator for text generation.
\newblock In \emph{Proceedings of the 2022 Conference on Empirical Methods in
  Natural Language Processing}, pages 2023--2038.

\end{thebibliography}

\appendix
\clearpage

\section{Rating Prediction Diversity}
\label{appendix:diversity}
To verify that \method could model/output more diverse rationales and ratings, we compare the standard deviation (SD) of the ratings predicted by different methods. For each story-aspect pair, we compute the SD of ratings (0-1 range) before averaging them into the final prediction. 

In \Cref{tab:main_baseline_results}, the SD of \textbf{\method (T5-3B)} (0.513) is much larger than the SD of \textbf{Mistral-7B-Instruct CoT SC Mean} (0.289) and \textbf{GPT 3.5 CoT SC Mean}  (0.33). In \Cref{tab:ablation}, the SD of \method (\textbf{T5 Small} $\mathbf{\rightarrow (\mathcal{K}_{\text{TextRank}}(e), e)}$ + $\mathbf{(s, a, \text{TextRank}(e)) \rightarrow}$ \textbf{DeBERTa-V3 Small} ) is 0.511, which is also much larger than 0.310 from $\mathbf{(a, e) \rightarrow}$ \textbf{DeBERTa-V3 Small} and 0.337 from $\mathbf{(s, a, e) \rightarrow}$ \textbf{DeBERTa-V3 Small}.

The experiment verify that keyword extraction indeed drastically improves the diversity of the predicted ratings and it also suggests that the models that has a larger Pearson's $\rho$ usually also has a larger SD (i.e., rating diversity).


\section{StoryER Dataset Analysis}
The StoryER dataset \citep{chen-etal-2022-storyer} extends the WritingPrompts \citep{fan-etal-2018-hierarchical} dataset, which consists of multiple writing prompts and corresponding human-written stories for those prompts, by adding ratings for ten \textit{aspects} that are picked by the authors from a given list of fixed aspects, along with \textit{comments} that justify the corresponding ratings given.

\begin{figure}[h!]
    \centering
    \includegraphics[width=0.4\linewidth]{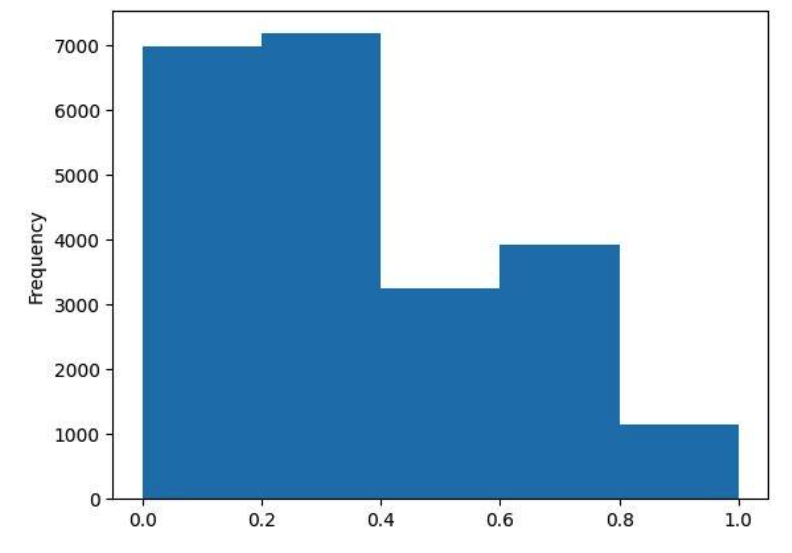}
    \caption{We plot the distribution of annotator provided ratings in the training set.}
    \label{fig:score-dist}
\end{figure}

\begin{figure}[h!]
    \centering
    \includegraphics[width=0.9\linewidth]{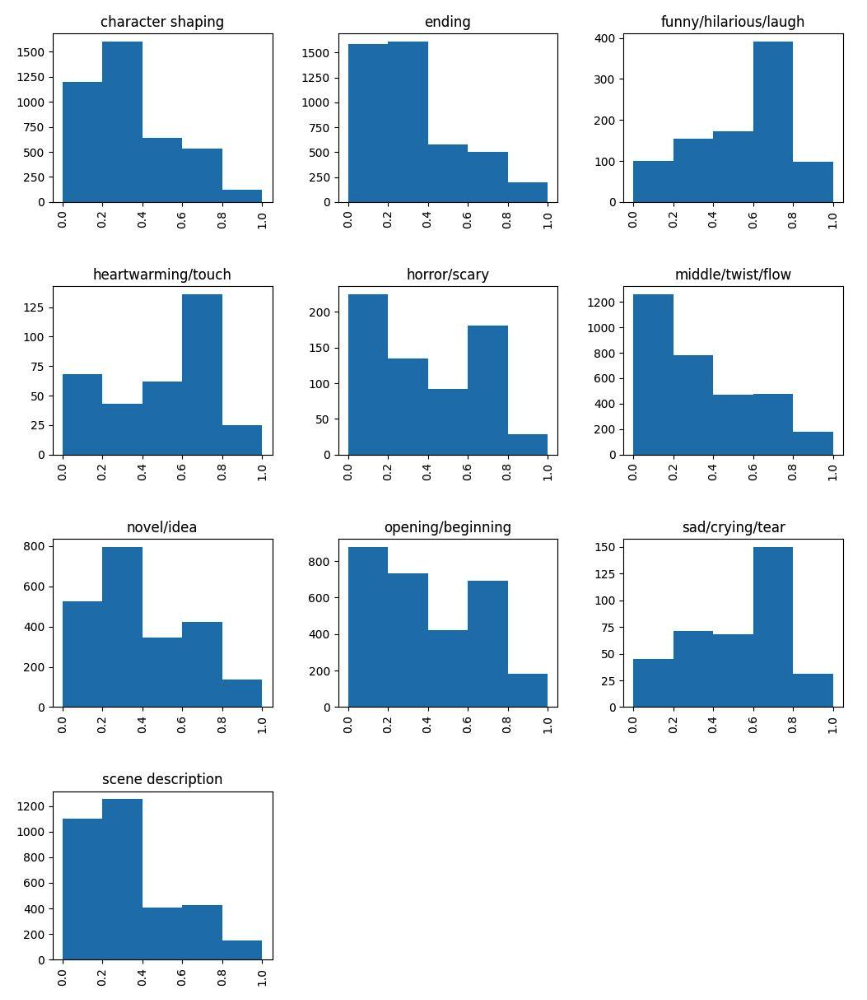}
    \caption{Distribution of annotator provided ratings across different aspects.}
    \label{fig:aspect-score-dist}
\end{figure}

Each of these aspects aims to highlight a separate semantic or literal aspect of the story -- for example, aspects can highlight the `ending' or `humour'-level of a story.
This is done by multiple annotators for every writing prompt + story pair, however the number of annotators, and actual aspects (out of ten) that are annotated for a story can vary.
\Cref{fig:score-dist} and \Cref{fig:aspect-score-dist} show the distribution of annotator provided ratings on the training set of the dataset.
\Cref{tab:dataset_details} and \Cref{tab:apx:storyer_examples} provide additional details of StoryER.

\begin{table}[h!]
\centering
\scalebox{0.85}{
\begin{tabular} {c|ccc}
\toprule
Split & Train & Dev & Test \\
\midrule
Number & 17982 & 4496 & 5631\\
\bottomrule
\end{tabular}
}
\caption{\textbf{Dataset details}: Since StoryER does not contain a validation set, we use the train set to create it. We partition the train set by unique writing prompts and split it into a train and validation set based on it.}
\label{tab:dataset_details}
\end{table}
\begin{table}[h!]
\centering
\begin{tabular}{lr}
\hline
\textbf{Aspect} & \textbf{Percentage} \\
\hline
Ending & 19.91\% \\
Character Shaping & 18.20\% \\
Scene Description & 14.81\% \\
Middle/Twist/Flow & 14.11\% \\
Opening/Beginning & 12.90\% \\
Novel/Idea & 9.90\% \\
Funny/Hilarious/Laugh & 4.08\% \\
Horror/Scary & 2.94\% \\
Sad/Crying/Tear & 1.62\% \\
Heartwarming/Touch & 1.48\% \\
\hline
\end{tabular}
\caption{\textbf{Percentage Distribution of Aspects in Training Set}: Given that not all aspects are annotated for all stories, there is an imbalance in the distribution of aspects.}
\label{tab:percentage-distribution}
\end{table}

\begin{table*}[h!]
\centering
\resizebox{\linewidth}{!}{
\begin{tabular}{{ p{0.35\linewidth} p{0.9\linewidth} p{0.2\linewidth} p{0.2\linewidth} }}
    \toprule
    \textbf{Writing Prompt} & \textbf{Story} & \textbf{Aspect, Score} & \textbf{Annotator Explanation}\\ 
    \midrule
    {The cure for death was discovered and it worked 99\% of the Earth's population. You are one of the 1\% and now 90 years later, you are the last mortal left on your deathbed. The World comes to see the last dying human.} & {The world didn't mourn. It was a celebration. Confetti, streamers, loud fireworks while I laid quietly on my deathbed.

    Death was dead. Long live life. Or so they thought.
    
    They didn't understand what I understood. It wasn't because the cure didn't work on me, no, it was because I *didn't* want the cure.
    
    Bodies rot. Minds decay. Death is a mercy to rid the world of these ugly things. Death wasn't the problem, humanity is.
    
    In time, they would realize it. They would remember my name, the last mortal to die, and cry for the ability to do so.
    
    Unfortunately for them, Death will never come in time.} & {novel/idea, 5/5} & {The story was really written for its tiny size, it ended and gave a powerful message to the reach of the humanity, i bet for them first 100 or 200 years will be wonderful, i don't know how they will control the population though} \\
    \midrule
    {You go to sleep on the night of your 25th birthday, only to wake up on your first day of 1st grade. You use your knowledge of the future to take advantage of the situation, and ball hard. However, when you come back to sleep the night of your 25th birthday, you wake up once again in 1st grade.} & {The clock ticks. I have one minute before I reach my silver year of life. I take this minute to reflect on my years.
    I was a very bratty child. I hated my teachers, as I thought that they were just other people in the world. I barely passed high school and took a couple weeks of college before I realized it wasn't what I was looking for in life. Since then, I had taken over my father's business in selling pools and spas as well as contracting. It was not a job I enjoyed, but it was one I had to do for my rent situation.
    3...2...1...
    11:42 PM on my birthday had passed. It was this day 25 years ago I had come out my mother's womb. Another year of a life that was just wasted. I had gone to sleep after this minute. Despite the momentous occasion, I still had a job to do early in the morning, and this customer was a particularly angry one.
    When I wake up, it is not the queen bed I have in my apartment, but the house I spent my early childhood in. Instead of the tall 6'3" body I had as an adult, I had the small body of a child. I look near my bed and see a face I had nearly forgotten. It was my old dog, Luna. She was already old when I was born and we were forced to put her down when I was merely 7 years old. I look at the calendar near my bed. It was about 19 years ago. I was 6 years old, about to go back to my first day of first grade.
    I realize something. First grade is when I changed from a curious child to a bratty child. Perhaps a higher power has sent me to fox my mistakes I have made.
    As I walk into class, I see many faces I had not seen in years. I look at my "beat friend" at this age, who grew up to be a crackhead. I look at my actual best friend, who looked just as snobbish as she described herself to be.
    Going home each day, I actually do my homework. I don't pay as much attention in class, as I had already learned this all in my old life. Over the years, I start making smarter decisions. Instead of joining a basketball league as a youth, I dedicate my time to writing stories, a dream I had in my teenagehood. The teachers view me as a prodigy who knows well past my age. I skip the 3rd grade due to my knowledge, but no more. Despite everything, I wasn't a prodigy in my past life, so I wasn't seen as the next Einstein.
    As I reach the puberty stages and a few years past that, I start attempting to care more for my body. Instead of having a mop for a head, I style my hair each day. By the time I am 15, I have a relationship with a friend from my past life, a stable one.
    Now, as I wait for my second 25th birthday, I sit back and realize that my life has changed. I managed to make my life better, but I cared little for others. Could I have done better. Of course I could. Would I want to start all over, of course not!
    3... 2... 1...
    11:42 has passed. I go to sleep.
    When I wake up, my body is once again too small and I wake up in an all to familiar bed. "It appears..." I whisper, "That you aren't done with me, yet."}. &  {character shaping, 2/5} & The author of this story was really unable to bring life to the identities and persona's of the characters in this story. Also they were no lively interactions between the characters. \\
        \midrule 
    {In the future criminals are thrown into a forest completely surrounded by city. Civilians hunt them in the forest. Police watch the forest edge for criminals, and kill them if seen leaving. You were falsely accused of murder and thrown into the forest with 4 other criminals.} & {They left us deep in the woods with nothing but our orange jumpsuits, our handcuffs, and each other. Fifteen minutes, they had told us. Fifteen minutes and the handcuffs would open. Fifteen minutes and the gates would open, letting the hunters in.

The others were talking. I ignored them. They were criminals, murderers. I was innocent.

I looked at my handcuffs. I knew how they worked. Each cuff had a tracking chip. When they sent the signal that opened the gates, the cuffs opened too. That was good information to have.

I rubbed my sternum. It was still sore. There was a tracking chip in me too, inside the bone. It tracked my position and heart rate. When I died, they would know it. If I tried to leave, they would see me. That was good information to have.

One of the others, Dan, was too loud. He broke my train of thought. I had to think. There was a way out, but I had to think.

"I won't be hunted! I won't! Not like some, some animal!" he shouted. "Some of them use dogs, you know! Better to just die now. If I make it to the edge, the guards will just shoot me. Better that way." He was rambling. He was frantic, manic.

"The edge is too far," I said. "You won't make it before they let the hunters in."

"Yeah, and what do you know? I heard you killed some kid. I done a lot of things, but I ain't never murdered no kid." He kept going. I ignored him. I hadn't killed anyone, at least not on purpose.

"Shut up, both of you," said Fat Mike. We called him Big Mike to his face. "We need to get ready. Need to make weapons," said Fat Mike.

"You want to fight guns with sticks?" Thin Mike scoffed. He was right.

Fat Mike was right too. They were coming to kill us. It was kill or be killed out here. I hadn't killed anyone, at least not on purpose. I had to think.

"Hey, where's Steve?" Fat Mike asked suddenly. I had noticed him slip off while the others were arguing, but I didn't say anything.

"He stole my idea!" proclaimed Dan. "He's headed to the edge. A man shouldn't be hunted. Better that way."

"I already told you, it's too far," I said.

"Shove it," Dan replied angrily. "Might as well try." He turned his back to me.

I slipped my hands over his head. The chain of my handcuffs pressed against his throat. I pulled as hard as I could. He struggled. "Better this way," I said. He struggled harder. I pulled harder. He stopped struggling. I let him fall. It had been easier than I thought it would be. That was good information to have.

The Mikes were quiet. I ignored them. My cuffs unlocked. I let them fall. They were coming to kill us. It was kill or be killed out here. I hadn't killed anyone, at least no one who wasn't asking for it. I had to think.}&  {scene description, 4/5} & {So actually the protagonist actually committed a crime and is not innocent, at least that's what was implied here "I hadn't killed anyone, at least not on purpose."}\\
    \bottomrule 
\end{tabular}
}
\caption{\textbf{StoryER Dataset:} We give some examples of how StoryER stories and aspects, as well as human annotator explanations look like.}
\label{tab:apx:storyer_examples}
\end{table*}


\section{\method Details}



\subsection{Training Parameters}

For all the LLM generations (on GPT 3.5, 4, LLaMa, and Mistral), we set a temperature of 1 and maximum token length of 1024. 

For training the rationalizer and scorer, we set the parameters as shown in \Cref{tab:training_details}.
The best checkpoints are chosen based on the lowest validation loss.

\begin{table}[h!]
\centering
\scalebox{0.75}{
\begin{tabular}{cc}
\toprule
\textbf{Config}&\textbf{Assignment}\\
\midrule
train batch size&4\\
eval batch size&4\\
seed&0\\
max epochs&25\\
learning rate&3e-5\\
learning scheduler&fixed\\
GPU&Quadro RTX 6000\\
Training time& 4 hours\\
\bottomrule
\end{tabular}
}
\caption{\textbf{Training Parameters}: Here we show the models we used and hyperparameters we used training.}
\label{tab:training_details}
\end{table}

\subsection{Human Performance Calculation} 

\begin{figure*}[h!]
    \centering
    \includegraphics[width=0.9\textwidth]{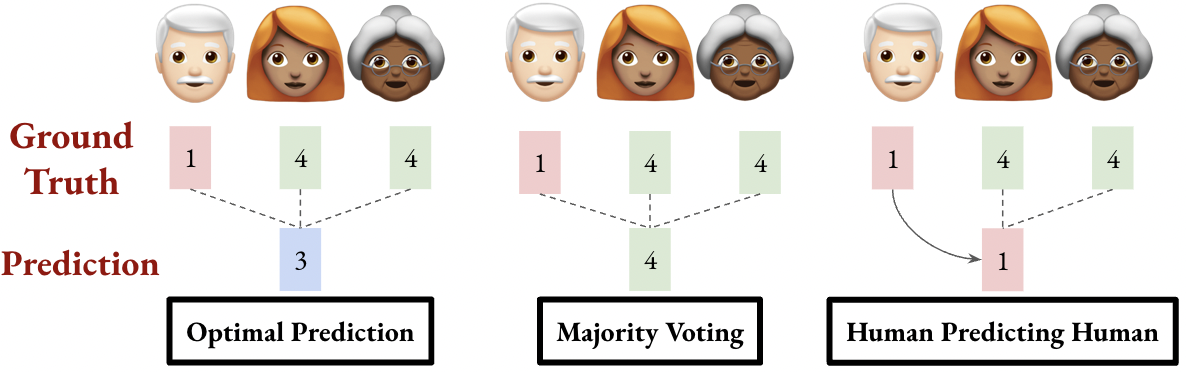}
    \caption{Different ways to calculate human performance - we use the \textit{human predicting human} setting.}
    \label{fig:human_perf}
\end{figure*}
We then calculate different variants of human performance that is estimated from the multiple-annotator annotations that the StoryER test set contains.
\autoref{fig:human_perf} contains a visual description of these variants.
Optimal Prediction and Majority Voting includes taking the mean and mode of the annotator predictions respectively as predictions.
However, they work under the assumption that ratings of all annotators are available at test time, which is not a realistic setting.
The Human Predicting Human variant randomly selects a rating from one annotator, and uses that as a prediction to estimate other annotators, which better represents the setting that our evaluation systems would fall into (assume the prediction from the system to be one `annotator' that tries to best approximate other annotators).

\subsection{Details about $\mathcal{K}(\cdot)$}
\label{sec:keyword_details}

For all of the keyword extractor methods, we set number of ngrams to be between one and three, so as to get a both keyword and keyphrases from the annotator explanations.
We extract the top ten keywords produced by these extractor.

\begin{table*}[h!]
\centering
\resizebox{\linewidth}{!}{
\begin{tabular}{{p{0.3\linewidth} p{0.35\linewidth} p{0.35\linewidth} p{0.5\linewidth} p{0.35\linewidth}}}
    \toprule
    \textbf{Aspect} & \textbf{Annotator Explanation} & \textbf{TextRank Keyword} & \textbf{Generated Keywords}& \textbf{Generated Explanation} \\ 
    \midrule
    {character shaping} & {The author didn't do a good job to portray their characters in this story. The author should've at least detailed his main character a little bit better.} & {'little bit', 'main character', 'characters', 'story', 'author'} & {naive cliche,main character,diverse characters,personalities,names,family,conversation} & {The two diverse characters were nicely written, their conversation with their families wasn't cliche and had personalities all their own, they didn't stand out in the crowd as much.} \\
    \midrule
    {heartwarming/touch} & {I would figure no matter the outcome when the kid came through the portal, even if your worst nightmare came out youd at least be cordial and make an attempt to be civil, not immediately come out swinging with the insults.} & {'worst nightmare', 'insults', 'attempt', 'kid', 'outcome', 'matter'} & {tame story,way,son,wife,decision,mom,man} & {I think this is a tame story because the man's decision to move in with his wife and son is pretty sweet. But the way he relates this is too shallow.} \\
    \midrule
    {ending} & {The ending didn't make any sense at all, the story was too boring and bland for my taste, i was keeping my wits together just to complete reading this story} & {wits,taste,story,sense,ending} & {toon science,story,divots,detailing,ending} & {The ending was kind of weird. I was expecting something about fixing the divots but there was no detailing or even detailing in the story.} \\    
    \bottomrule 
\end{tabular}
}
\caption{\textbf{Example Generations:} We give some examples of how StoryER annotator explanations and extracted keywords look, along with generated keywords and explanations.}
\label{tab:apx:keyword_examples}
\end{table*}

\subsection{Human Evaluation}
\label{sec:apx:human_eval}

\begin{figure*}[h!]
    \centering
    \includegraphics[width=0.9\textwidth]{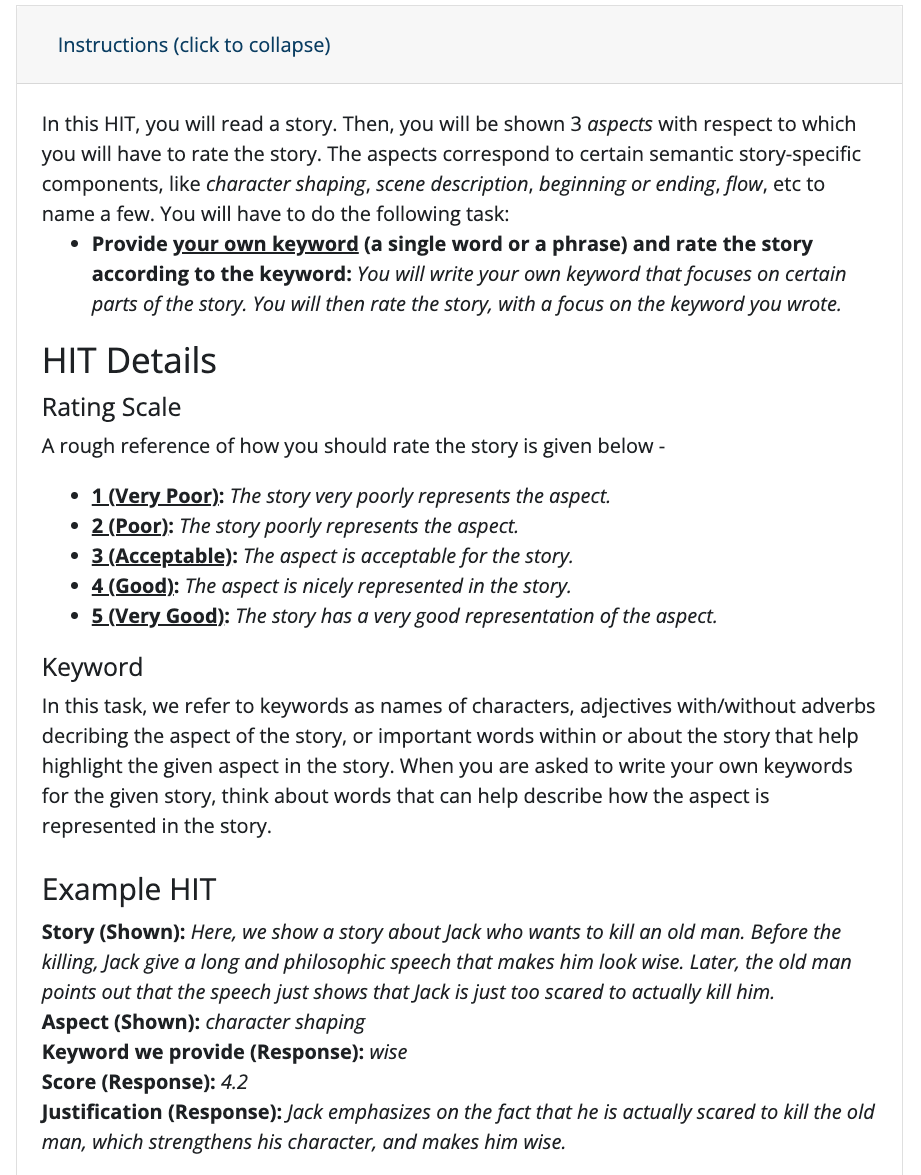}
    \caption{Instructions provided to turkers.}
    \label{fig:turk_instruction}
\end{figure*}

\begin{figure*}[h!]
    \centering
    \includegraphics[width=0.9\textwidth]{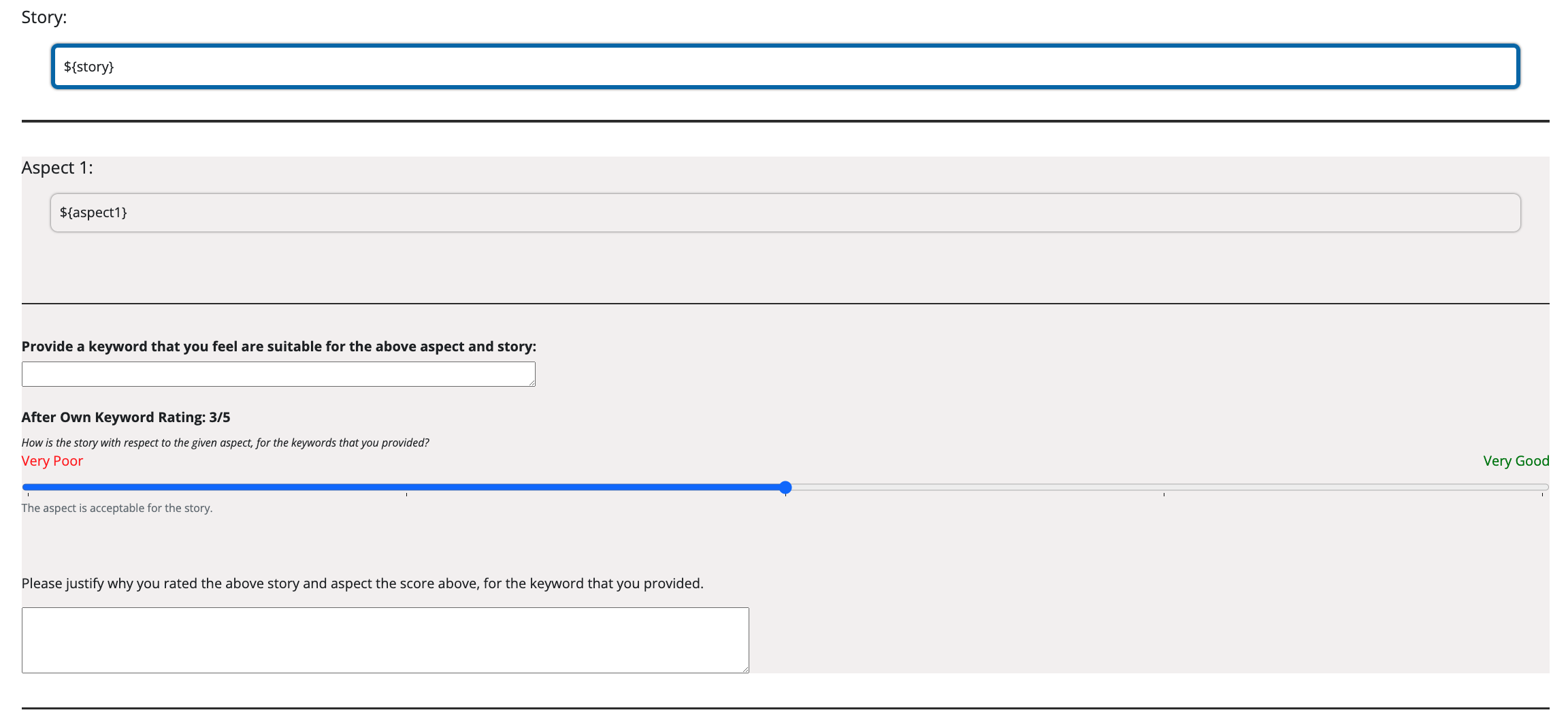}
    \caption{Actual task given to turkers.}
    \label{fig:turk_task}
\end{figure*}

All our crowdworkers are from countries where English is the primary language.
For all our human studies, the task is setup in a manner that ensure that the annotators receive compensation that is above minimum wage.
Turkers were also chosen using extensive qualifications, where they had prior story reading and rating experience. 
We provide the task shown to turkers in \Cref{fig:turk_instruction,fig:turk_task}.

\end{document}